\documentclass[lettersize,journal]{IEEEtran}
\usepackage{cite}
\usepackage{amsmath,amsfonts,amssymb}
\usepackage{algorithm}
\usepackage{algorithmicx}
\usepackage{algpseudocode}
\usepackage{array}
\usepackage{textcomp}
\usepackage{url}
\usepackage{graphicx}

\usepackage{booktabs}       % 表格三线
\usepackage{tabularx}       % 表格宽度
\usepackage{multirow}       % 表格合并
\usepackage{makecell}       % 提供上下居中功能

\usepackage{bm}             % 数学粗体

\usepackage[dvipsnames]{xcolor}
\usepackage[colorlinks=true,linkcolor=Cerulean,citecolor=Cerulean,urlcolor=Cerulean]{hyperref}
\usepackage{orcidlink}
\usepackage{cleveref}       % 引用
\crefformat{figure}{Fig.~#2#1#3}            % 修改figure的引用前缀为 "Fig."
\crefformat{table}{Table~#2#1#3}            % 修改table的引用前缀为 "Table"
\crefformat{algorithm}{Algorithm~#2#1#3}    % 修改algorithm的引用前缀为 "Algorithm"
\crefformat{section}{#2Section~#1#3}        % 修改section的引用前缀为 "Section"
\crefformat{subsection}{#2Section~#1#3}     % 修改subsection的引用前缀为 "Section"

\usepackage{placeins}       % 浮动设置

\begin{document}
% 用于专用模型适配的开放式场景推理
\title{Open-Ended Scenario Reasoning for Specialist Model Adaptation}

\author{
  Youcheng Zong\(^{\orcidlink{0009-0008-6795-8412}}\),~\IEEEmembership{Student~Member,~IEEE},
  Runda Jia\(^{\orcidlink{0000-0002-8586-243X}}\),
  Ranmeng Lin\(^{\orcidlink{0009-0006-3829-4472}}\),
  Mingxuan Ren\(^{\orcidlink{0009-0001-0172-5914}}\),
  \\ and Dakuo He\(^{\orcidlink{0000-0001-8303-529X}}\)
  \thanks{This work was supported by the Fundamental Research Funds for the Central Universities, China (N26GFZ006). \textit{(Corresponding author: Runda Jia.)}}
  \thanks{Youcheng Zong, Runda Jia, Ranmeng Lin, Mingxuan Ren, and Dakuo He are with the College of Information Science and Engineering, Northeastern University, Shenyang 110004, China (e-mail: youchengzong@stumail.neu.edu.cn; jiarunda@ise.neu.edu.cn; linrm@mails.neu.edu.cn; renmx@mails.neu.edu.cn; hedakuo@ise.neu.edu.cn).}
  \thanks{The source code is available at \href{https://github.com/mituan-ai/ROAM_open}{https://github.com/mituan-ai/ROAM\_open}.}
}

\maketitle

\begin{abstract}
  % 过程工业积累了大量经过验证的专用模型，但传感器漂移、来料变化和工况切换会使这些模型在新场景下系统性退化。重新采集标注数据并重训练代价高昂，而继续使用原始模型则会带来持续偏差。
  Process industries have accumulated validated specialist models, yet sensor drift, feedstock variation, and regime switching cause these models to degrade systematically in new scenarios. Collecting new labeled data and retraining is costly, while continuing with the original model incurs persistent bias.
  % 现有适配方法需要修改模型参数和充足的标注数据，难以在已部署系统上快速响应。将 LLM 用作直接预测器则面临幻觉和不可控输出的风险，也无法融合现场的非结构化场景知识。
  Existing adaptation methods require modifying model parameters with sufficient labeled data, making rapid response on deployed systems difficult. Using LLMs as direct predictors risks hallucinations and uncontrollable outputs. Such predictors also cannot incorporate unstructured scenario knowledge from the field.
  % 为解决上述问题，本文提出面向专家模型的推理驱动开放式自适应 (ROAM)，一种利用 LLM 世界知识与推理能力使冻结模型无需重训练即可适应未见场景的框架。ROAM 将所有校正约束在低维语义可解释的 latent 空间中，在统一概率框架下融合 LLM 生成的场景判断与在线观测。风险约束机制会在 LLM 证据不可靠或场景剧变时抑制校正，并在证据不足时回退到原始冻结模型。
  To address these limitations, this article proposes Reasoning-Driven Open Adaptation for Specialist Models (ROAM), a framework that uses LLM world knowledge and reasoning to adapt frozen specialist models to unseen scenarios without retraining. ROAM confines all corrections to a low-dimensional, semantically interpretable latent space. LLM-generated scenario judgments and online observations are fused under a unified probabilistic framework. A risk-constrained mechanism suppresses corrections under unreliable LLM evidence or abrupt scenario shifts and falls back to the original frozen model when evidence is insufficient.
  % 在矿物浓密过程和公开的 IndPenSim 青霉素发酵数据集上的实验表明，ROAM 在隐藏偏移等主要偏移场景中将 MAE 降低超过 20\%，仅需 839 个额外参数和不到 0.02\,ms 的每步开销。这些结果表明，LLM 推理能力可以转化为已部署工业模型的保守适配信号。
  Experiments on a mineral thickening process and the public IndPenSim penicillin fermentation dataset show that ROAM reduces MAE by over 20\% in major shift settings such as hidden shifts with only 839 additional parameters and under 0.02\,ms per-step overhead. These results indicate that LLM reasoning can be turned into a conservative adaptation signal for industrial models already in service.
\end{abstract}

% 给从业者的说明
\section*{Note to Practitioners}
% 本文面向过程工厂中的一个实际问题：质量、浓度或产率等关键变量通常不能连续测量，因此自动化系统依赖软测量模型给出在线估计。这些模型在投运前已经校验，但传感器漂移、原料性质变化、设备维护或控制策略切换之后，估计值可能缓慢偏离真实过程。重新采样、化验和重训练往往需要数小时到数天，在此期间有偏估计会影响控制、报警和排产决策。本文提出的 ROAM 可作为现有软测量模型之外的一层适配器使用。它读取班组日志、检修记录和新运行区间开始时的少量过程信号，判断当前偏差更可能来自测量偏置、比例变化、负荷变化、动态响应变化还是输出关系变化。随后，ROAM 只对模型输出施加受限校正；当记录不清楚、信号矛盾或新状态离已知运行区域太远时，它会减小校正或回到原模型。这样做的实际价值是保留已验证的模型资产，同时在异常或新工况早期减少持续偏差，从而提高质量估计的可靠性并降低紧急重训练的需求。该方法目前仍需要可用的现场记录，并假设一个运行区间内主导工况不会频繁变化。后续工作应接入报警序列、图像巡检、维护系统和操作员反馈，并在更长时间的生产闭环中验证。除浓密脱水和发酵过程外，同一思路也可用于能源负荷预测、设备健康监测和其他需要保留既有模型认证的自动化应用。
This paper is motivated by a practical problem in process plants: key variables such as quality, concentration, or yield are often not measured continuously, so automation systems rely on soft sensors for online estimates. These models are validated before deployment, but their estimates can slowly drift from the real process after sensor drift, feedstock changes, equipment maintenance, or control-policy switching. New sampling, laboratory analysis, and retraining may take hours or days. During this period, biased estimates can affect control, alarms, and production scheduling. ROAM is intended as an adaptation layer outside the existing soft sensor. It reads shift logs, maintenance records, and a small set of process signals at the start of a new operating interval, then judges whether the current error is more likely due to measurement bias, scaling change, load change, dynamic-response change, or output-relation change. ROAM then applies only a bounded correction to the model output. When records are unclear, signals conflict, or the new state is too far from known operation, it reduces the correction or returns to the original model. The practical value is to keep validated model assets in service while reducing persistent bias early in abnormal or new conditions, improving the reliability of quality estimates and reducing the need for urgent retraining. The current approach still needs usable field records and assumes that the dominant operating condition does not change frequently within one interval. Future work should connect alarm sequences, inspection images, maintenance systems, and operator feedback, and test the method in longer closed-loop production. Beyond thickening dewatering and fermentation, the same idea may support energy-load forecasting, equipment-health monitoring, and other automation applications where existing model certification must be preserved.

% 关键词
\begin{IEEEkeywords}
  % 大语言模型（LLM），贝叶斯推断，在线适配，过程工业，软测量。
  large language model (LLM), Bayesian inference, online adaptation, process industries, soft sensing.
\end{IEEEkeywords}

%<WPRF:BEGIN Intro>
% 引言
\section{Introduction}\label{sec:1}

% 矿物加工、石油化工和生物制药等过程工业，已经积累了经过验证的 data-driven specialist models，用于关键质量变量的软测量。相关研究已经处理了动态建模、基础模型迁移、缺失数据、部分标注和物理一致性等关键建模约束。这些模型在训练工况下表现良好，但一旦传感器漂移、来料变化、工况切换或设备老化引入分布偏移，它们就会系统性退化。
Process industries including mineral processing, petrochemicals, and biopharmaceuticals have accumulated validated data-driven specialist models for soft sensing of critical quality variables~\cite{kadlec2009softsensor,yuan2020lstm,geng2022transformer,su2025zero}. Related studies have addressed key modeling constraints such as dynamic modeling, foundation-model transfer, missing data, partial labels, and physical consistency~\cite{yao2022figan,shi2025semisupervised,wang2026spatiotemporal}. These models perform well under their training conditions but degrade systematically once sensor drift, feedstock variation, regime switching, or equipment aging introduces distribution shifts~\cite{sun2021survey}.
% 例如，在矿物浓密脱水中，底流浓度的软测量直接决定脱水效率和尾矿库安全。来料性质、沉降动力学和观测链路等偏移来源会同时作用在同一个模型上。重新标注和重训练成本高、工程窗口窄，而继续使用原始模型则会带来系统性偏差。这一困境在过程工业中普遍存在。
In mineral thickening dewatering, for instance, soft sensing of underflow concentration directly governs dewatering efficiency and tailings storage safety. Shift sources including feedstock properties, settling dynamics, and observation pathways act on the same model simultaneously. Relabeling and retraining are costly with narrow engineering windows, while continuing with the original model incurs systematic bias. This dilemma is common across process industries~\cite{zong2025hybridgrid}.

% 现场工程师通常掌握矿石品位下降、药剂配方调整等场景级知识，并记录在维护日志和交接报告中，但已部署的模型没有机制利用这些证据。现有适配策略在三个层面上回避了这一问题。迁移学习、域适应、元学习等参数更新方法假设模型参数可修改，违背冻结约束且适配信号完全来自数据分布。基于 LLM 的方法将 LLM 置于预测环路内部，缺乏不确定性控制。事后可解释性方法的诊断与适配决策完全解耦。
Field engineers often possess scenario-level knowledge such as ore-grade decline or reagent-formula changes, recorded in maintenance logs and shift reports, yet the deployed model has no mechanism to consume such evidence. Existing adaptation strategies sidestep this problem at three levels. Parameter-update methods such as transfer learning, domain adaptation, and meta-learning assume modifiable parameters, violating the frozen constraint and drawing signals solely from data distributions~\cite{chai2022transfer,zhang2024subdomain,finn2017maml}. LLM-based methods place the LLM inside the prediction loop without uncertainty control~\cite{chen2025dkilm,chen2026llmsa}. Post-hoc interpretability methods decouple diagnosis from adaptation entirely~\cite{jang2025xaifd}.
% 这三个盲区指向同一个根本问题：开放证据鸿沟。场景知识以非结构化、不确定的开放文本存在，而安全的模型适配需要结构化、带不确定性量化的保守校正信号。弥合这一鸿沟需要一个范式转变，从修改模型参数到更新关于场景状态的信念。
These blind spots converge on one root cause: the open-ended evidence gap. Scenario knowledge exists as unstructured, uncertain text, yet safe adaptation demands structured, uncertainty-quantified corrections. Bridging this gap requires a paradigm shift from updating model parameters to updating beliefs about the scenario state.

\begin{figure}[!t]
  \centering
  \includegraphics[width=88mm]{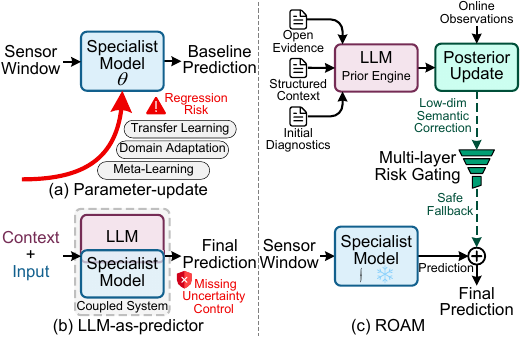}
  % 工业专用模型的适配范式。(a) 参数更新方法直接修改 $\theta$，会对已验证行为带来回归风险，而且没有开放式文本证据的入口。(b) LLM-as-predictor 方法将 LLM 嵌入预测环路，但缺乏显式不确定性控制。(c) ROAM 冻结 $\theta$，只将 LLM 用作环外先验引擎。适配被限制在低维语义 latent 校正中，通过贝叶斯后验更新融合，并经过多层风险门控过滤。证据不足时，系统回退到基线预测。
  \caption{Adaptation paradigms for industrial specialist models. (a)~Parameter-update methods modify~$\theta$ directly, risking regression on validated behavior with no entry point for open-ended textual evidence. (b)~LLM-as-predictor approaches embed the LLM in the prediction loop without explicit uncertainty control. (c)~ROAM freezes~$\theta$ and uses the LLM only as an out-of-loop prior engine. Adaptation is confined to a low-dimensional semantic latent correction, fused by Bayesian posterior updating and filtered through multi-layer risk gating. The system falls back to the baseline prediction when evidence is insufficient.}
  \label{fig:paradigm_comparison}
\end{figure}

% 如 \cref{fig:paradigm_comparison}(c) 所示，我们提出面向专家模型的推理驱动开放式自适应（ROAM），将适配重新表述为信念更新问题。ROAM 将所有适配自由度约束在语义锚定的低维 latent 空间中。LLM 仅作为外部结构化先验引擎，产生带校准不确定性的信念判断，而不进入预测管线。语义先验与异质在线观测在同一概率框架中通过贝叶斯后验更新融合。整个适配链路被设计为保守的：当证据稀缺时，系统回退到原始冻结预测。这种解耦设计使 LLM 和 specialist model 可以独立升级，而无需相互重新验证。
As illustrated in \cref{fig:paradigm_comparison}(c), we propose Reasoning-Driven Open Adaptation for Specialist Models (ROAM), which recasts adaptation as a belief-update problem. ROAM constrains all adaptation degrees of freedom to a semantically anchored, low-dimensional latent space. The LLM serves solely as an external structured prior engine that produces belief judgments with calibrated uncertainty, without entering the prediction pipeline. Semantic priors and heterogeneous online observations are fused through Bayesian posterior updating within a single probabilistic framework. The entire adaptation chain is designed to be conservative: the system falls back to the original frozen prediction when evidence is scarce. This decoupled design allows independent upgrades of the LLM and the specialist model without mutual revalidation.
% 主要贡献如下。
The main contributions are as follows.
\begin{itemize}
  \item
        % 我们提出 ROAM，它利用 LLM 的世界知识与推理能力，使冻结模型无需重训练即可适应未见场景。
        We propose ROAM, a framework that uses LLM world knowledge and reasoning to enable frozen models to adapt to unseen scenarios without retraining.
  \item
        % 我们引入风险约束机制，在 LLM 幻觉、证据不足或场景剧变时抑制不可靠校正，并支持回退到原始冻结模型。
        We introduce a risk-constrained mechanism that suppresses unreliable corrections under LLM hallucinations, insufficient evidence, or abrupt scenario shifts, and supports fallback to the original frozen model.
  \item
        % 在浓密脱水过程和青霉素发酵基准上，ROAM 在主要偏移场景中带来跨主干的显著性能增益，同时只增加约 0.02\,ms 的额外推理开销。
        On both a thickening dewatering process and a penicillin fermentation benchmark, ROAM yields substantial cross-backbone gains in major shift settings with only $\sim$0.02\,ms of additional inference overhead.
\end{itemize}

% 本文余下部分安排如下。\cref{sec:2} 回顾相关工作，\cref{sec:3} 介绍所提出的 ROAM 框架，\cref{sec:4} 报告实验结果，\cref{sec:5} 对本文进行总结。
The remainder of this paper is organized as follows. \cref{sec:2} reviews related work, \cref{sec:3} presents the proposed ROAM framework, \cref{sec:4} reports experimental results, and \cref{sec:5} concludes this article.
%<WPRF:END Intro>

%<WPRF:BEGIN RelatedWork>
% 相关工作
\section{Related Work}\label{sec:2}

% 参数更新适配
\subsection{Parameter-update adaptation}
% 迁移学习通过微调参数适配目标域，域适应通过对齐特征分布减小跨域差异。跨域工业诊断进一步利用原型或可迁移特征缓解域间差异。
Transfer learning adapts to target domains by fine-tuning parameters~\cite{chai2022transfer}, and domain adaptation aligns feature distributions to reduce cross-domain discrepancy~\cite{zhang2024subdomain}. Cross-domain industrial diagnosis further uses prototypes or transferable features to reduce domain gaps~\cite{chai2022faultprototypical}.
% 元学习通过 episodic training 获得快速适配能力，而在线学习与测试时适配会在部署阶段持续更新参数。
Meta-learning acquires rapid adaptation capability through episodic training~\cite{zong2025metacontrastive}, and online learning together with test-time adaptation continuously updates parameters at deployment~\cite{wang2021tent}.
% 这些方法的有效性建立在一个前提上：模型参数是适配的合法自由度。在工业部署中，specialist model 在验证后被冻结，参数修改使认证失效并引入回归风险。即使参数可更新，适配信号也完全来自数据分布，无法利用开放文本形式的场景知识。
These methods rest on the premise that model parameters are legitimate degrees of freedom for adaptation. In industrial deployments, however, specialist models are frozen after validation; parameter modification invalidates certification and introduces regression risk~\cite{sun2021survey}. Even if parameters were updatable, adaptation signals come entirely from data distributions, with no mechanism to exploit scenario knowledge in textual form.

% LLM 在工业预测中的应用
\subsection{LLMs in industrial prediction}
% 近期研究主要从三种角度将 LLM 引入工业预测与决策。第一类将 LLM 作为预测模型的一部分，通过微调、适配器或跨模态对齐，把文本因素或语义嵌入接入电池健康估计、负荷预测、非平稳时序预测和故障诊断任务~\cite{chen2025dkilm,chen2026llmsa,chu2026a2ra,han2026zeroshot}。第二类将 LLM 作为离线知识源，为图神经网络提供物理先验或构建工业知识图谱~\cite{yao2026causalllm,liu2024kgllm}。第三类面向人机协同决策支持，使 LLM 在冶金等复杂工业过程中整合现场知识并生成操作建议~\cite{zong2026llm}。与此同时，近期分析也指出，LLM 在时序预测中相对简单基线的真实增益仍需谨慎评估~\cite{merrill2024llmts}。
Recent studies introduce LLMs into industrial prediction and decision support from three perspectives. The first treats the LLM as part of the prediction model, using fine-tuning, adapters, or cross-modal alignment to connect textual factors or semantic embeddings with battery health estimation, load forecasting, nonstationary time-series prediction, and fault diagnosis~\cite{chen2025dkilm,chen2026llmsa,chu2026a2ra,han2026zeroshot}. The second uses LLMs as offline knowledge sources, providing physics-informed priors for graph neural networks or constructing industrial knowledge graphs~\cite{yao2026causalllm,liu2024kgllm}. The third focuses on human-AI collaborative decision support, where LLMs integrate field knowledge and generate operational suggestions for complex industrial processes such as metallurgy~\cite{zong2026llm}. At the same time, recent analysis suggests that the practical gain of LLMs over simpler baselines in time-series forecasting should be assessed carefully~\cite{merrill2024llmts}.
% 结合多模态证据和动态变量依赖的工业诊断工作进一步表明，开放证据可以支持现场状态判断，但这些证据通常仍服务于诊断或分类，而不是冻结预测模型的保守校正。上述方向表明 LLM 可以提供语义、知识和决策层面的补充信息，但它们仍没有解决安全适配所需的结构化信念问题。预测管线内方法通常缺乏显式不确定性控制；离线知识注入方法主要在训练阶段使用静态知识；人机协同系统强调操作建议，但不直接给出对冻结预测模型的可审计校正机制。因此，LLM 的语义判断（如"传感器发生了漂移"）仍需要被转化为带置信度和边界的结构化信念。否则，LLM 与 specialist model 的耦合会使任一组件升级都需要重新验证整个系统。
Industrial diagnosis work that fuses multimodal evidence and dynamic variable dependencies further shows that open-ended evidence can support field-state judgment, but such evidence usually supports diagnosis or classification rather than conservative correction of frozen predictive models~\cite{wang2026adaptive,ma2026dynamic}. These directions show that LLMs can provide complementary semantic, knowledge-level, and decision-level information, but they do not solve the structured-belief problem required for safe adaptation. Prediction-in-the-loop methods usually lack explicit uncertainty control. Offline knowledge-injection methods mainly use static knowledge during training. Human-AI collaborative systems emphasize operational suggestions, but do not directly provide an auditable correction mechanism for frozen predictive models. Therefore, an LLM's semantic judgment (e.g., ``sensor drift has occurred'') must still be converted into a structured belief with confidence and boundaries. Otherwise, coupling the LLM with the specialist model means that upgrading either component requires revalidating the whole system.

% 可解释性、不确定性与适配安全性
\subsection{Interpretability, uncertainty, and safety}
% XAI-FD 结合对抗自编码器与 SHAP 做事后归因，CDGNN 通过因果解耦图神经网络在训练阶段学习因果与偏置子结构的分离。这些方法将可解释性视为事后分析层，解释不影响适配决策，也没有闭环机制将诊断转化为模型校正。
XAI-FD combines adversarial autoencoders with SHAP for post-hoc attribution~\cite{jang2025xaifd}, and CDGNN separates causal and bias substructures through a causal disentangled graph neural network~\cite{liu2025cdgnn}. These methods treat interpretability as a post-hoc analysis layer: the explanation does not influence adaptation, and no closed-loop mechanism translates diagnosis into correction.
% 贝叶斯方法、概率集成和统计特征学习为不确定性量化与工业过程在线故障诊断提供了原则性工具。然而，现有贝叶斯方法主要量化模型参数或预测输出中的不确定性，而非适配过程本身。现有方法也很少将安全机制纳入适配设计，在证据不足或矛盾时缺少自动抑制校正的能力。
Bayesian methods, probabilistic ensembles, and statistical feature learning provide principled tools for uncertainty quantification and online fault diagnosis in industrial processes~\cite{chen2024bayesiangnn,jiang2023bayesian,yu2019online,li2025twofold}. Existing Bayesian approaches, however, quantify uncertainty in model parameters or outputs, not in the adaptation process itself. Few methods incorporate safety into the adaptation design or automatically suppress corrections when evidence is insufficient or contradictory.
% ROAM 是最早一批将 LLM 放在预测环路之外作为结构化先验引擎、在部署阶段通过贝叶斯后验更新执行在线证据融合、并通过多层风险门控确保安全回退的框架之一。\cref{sec:3} 详述了该框架。
ROAM is among the first frameworks that place the LLM outside the prediction loop as a structured prior engine, perform online evidence fusion through Bayesian posterior updating at deployment, and ensure safe fallback via multi-layer risk gating. \cref{sec:3} details the framework.
%<WPRF:END RelatedWork>

\section{Method}\label{sec:3}

\subsection{Overview}\label{sec:3-1}

% ROAM 由离线阶段和在线阶段组成。离线阶段在训练场景上学习一组低容量线性映射，建立从主干隐藏态到五维语义 latent 空间的适配流形（\cref{sec:3-3}）。在线阶段对每个新 episode 依次执行三个步骤：语义先验构造（\cref{sec:3-2}）、结构化后验更新（\cref{sec:3-4}）以及带风险门控的输出校正（\cref{sec:3-5}）。
ROAM consists of an offline phase and an online phase. The offline phase learns a set of low-capacity linear mappings on training scenarios, establishing an adaptation manifold from backbone hidden states to a five-dimensional semantic latent space (\cref{sec:3-3}). The online phase executes three stages for each new episode: semantic prior construction (\cref{sec:3-2}), structured posterior updating (\cref{sec:3-4}), and output correction with risk gating (\cref{sec:3-5}).

% 设 $f_\theta$ 为一个已在有限训练场景上离线训练并冻结的序列回归器（specialist model），其参数 $\theta$ 在部署后不再更新。$f_\theta$ 以滑动窗口 $\mathbf{x}_{t-W+1:t}$（$W$ 为窗口长度）为输入，输出标量预测 $\hat{y}_t^0$ 和内部隐藏态 $\mathbf{h}_t$。ROAM 在不修改 $\theta$ 的前提下，利用开放式场景证据和在线观测估计并补偿残余环境效应。
Let $f_\theta$ be a sequence regressor (specialist model) trained offline on a limited set of scenarios and frozen at deployment, with parameters $\theta$ that are never updated after deployment. $f_\theta$ takes a sliding window $\mathbf{x}_{t-W+1:t}$ ($W$ is the window length) as input and produces a scalar prediction $\hat{y}_t^0$ and an internal hidden state $\mathbf{h}_t$. ROAM estimates and compensates for residual environmental effects using open-ended scenario evidence and online observations, without modifying $\theta$.

% 我们将部署时间线划分为若干 episode。一个 episode $r$ 是一段连续运行区间，在此期间场景状态近似恒定。episode 边界由外部事件（如工况切换、维护干预或操作员判断）触发，新 episode 开始时后验重置为新构造的先验。对每个 episode $r$，我们引入一个低维的纠偏潜变量（corrective latent）$\mathbf{z}_r \in \mathbb{R}^5$，其五个维度分别锚定到工业现场最常见的失配来源。\cref{tab:semantic_axes} 列出了各轴的语义定义及其工业对应。
We partition the deployment timeline into episodes. An episode $r$ is a contiguous operating interval during which the scenario state is approximately stationary. Episode boundaries are triggered by external events (e.g., operating-condition switching, maintenance intervention, or operator judgment), and the posterior is reset to a freshly constructed prior at the start of each new episode. For each episode $r$, we introduce a low-dimensional corrective latent $\mathbf{z}_r \in \mathbb{R}^5$, with each dimension anchored to a common source of industrial mismatch. \cref{tab:semantic_axes} lists the semantic definition and industrial correspondence of each axis.

\begin{table}[t]
  \centering
  % 5-D 纠偏 latent 空间的语义轴。每个轴都对应工业部署中的一种常见失配来源。
  \caption{Semantic axes of the 5-D corrective latent space. Each axis corresponds to a common source of mismatch in industrial deployment.}
  \label{tab:semantic_axes}
  \setlength{\tabcolsep}{2pt}
  \begin{tabularx}{\linewidth}{llXX}
    \toprule
    Axis     & Symbol & Semantic                     & Industrial Correspondence                                      \\
    \midrule
    bias     & $z_1$  & fixed offset in observations & sensor zero-point drift, static measurement bias               \\
    scale    & $z_2$  & proportional distortion      & sensor range drift, observation scaling change                 \\
    load     & $z_3$  & hidden load variation        & upstream feed property change, latent exogenous disturbance    \\
    dynamics & $z_4$  & response rate change         & process inertia variation, post-switching temporal dynamics    \\
    readout  & $z_5$  & output mapping drift         & baseline readout shift, hidden-state-to-target relation change \\
    \bottomrule
  \end{tabularx}
\end{table}

% 此外，系统维护一个观测异常分数 $\tau_r \in [0,1]$（不属于 $\mathbf{z}_r$ 的维度，而是 episode 级的门控变量），用于门控观测相关校正的强度。$\tau_r$ 越高表示当前观测链路异常程度越大，观测侧校正将被越强地抑制。
In addition, the system maintains an observation anomaly score $\tau_r \in [0,1]$ (not a dimension of $\mathbf{z}_r$ but an episode-level gating variable) that gates the strength of observation-related corrections. A higher $\tau_r$ indicates greater anomaly in the current observation pathway, leading to stronger suppression of observation-side corrections.

% 综上，所有适配自由度被约束在语义锚定的低维流形（$\mathbf{z}_r$ 和 $\tau_r$）上，以降低过度适配扰动冻结主干既有行为的风险。
Together, all adaptation degrees of freedom are constrained to a semantically anchored, low-dimensional manifold ($\mathbf{z}_r$ and $\tau_r$), reducing the risk that over-adaptation disturbs the frozen backbone's established behavior.

\begin{figure*}[t]
  \centering
  \includegraphics[width=179mm]{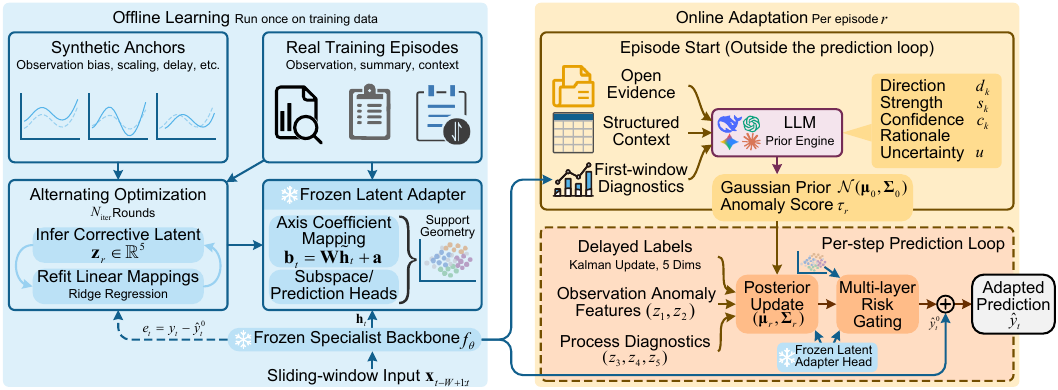}
  % ROAM 框架概览。左半部分显示离线阶段，它从训练 episode 残差中合成锚点，并交替执行 corrective-latent 推断与线性映射重拟合，以产出冻结的 latent adapter 和 support geometry。
  % 右半部分显示在线阶段。它根据 LLM 导出的语义判断构造高斯先验，通过贝叶斯后验更新融合延迟标签、观测异常特征和过程诊断，并应用多层风险门控以得到适配后的预测。
  \caption{Overview of the ROAM framework. The left half shows the offline phase, which synthesizes anchors from training-episode residuals and alternates corrective-latent inference with linear-mapping refitting to produce a frozen latent adapter and support geometry. The right half shows the online phase. It constructs a Gaussian prior from LLM-derived semantic judgments, fuses delayed labels, observation anomaly features, and process diagnostics via Bayesian posterior updating, and applies multi-layer risk gating to yield the adapted prediction.}
  \label{fig:roam_framework}
\end{figure*}

\subsection{Semantic Prior Construction via LLM Reasoning}\label{sec:3-2}

% ROAM 使用大语言模型（LLM）作为先验引擎，将 episode 开始时可获得的开放式证据转换为五个语义轴上的结构化先验判断。先验引擎不参与预测管线，它只提供先验信念。
ROAM uses a large language model (LLM) as a prior engine to convert open-ended evidence available at the episode start into structured prior judgments on the five semantic axes. The prior engine does not participate in the prediction pipeline---it only provides prior beliefs.

% 先验引擎的输入由三部分组成，均遵循时间合法性原则（不使用未来信息）。(1) 开放证据：来自操作记录、检修记录和上游说明的场景级文本。(2) Episode 起点结构化上下文：episode 边界处时间合法的任务相关外生上下文。(3) 首窗诊断：对冻结主干在首个完整窗口上做一次前向推理（此时无标签可用），提取数值异常信号，如特征漂移分数和隐藏态统计量偏移。
The prior engine input has three temporally legal parts (no future information is used). (1)~Open evidence: scenario-level text from operation logs, maintenance records, and upstream descriptions. (2)~Episode-start structured context: task-specific exogenous context available at the episode boundary. (3)~First-window diagnostics: a single forward pass of the frozen backbone on the first complete window before any labels arrive, extracting numerically defined anomaly signals such as feature-drift scores and hidden-state statistics shifts.

% 先验引擎对六个语义轴（五个纠偏轴加 trust）各输出一组结构化判断：方向 $d_k \in [-1, 1]$、强度 $s_k \in [0, 1]$、置信度 $c_k \in [0, 1]$ 和自然语言依据，以及一个全局不确定性 $u \in [0, 1]$。
The prior engine outputs a structured judgment for each of the six semantic axes (five corrective axes plus trust): direction $d_k \in [-1, 1]$, strength $s_k \in [0, 1]$, confidence $c_k \in [0, 1]$, a natural-language rationale, and a global uncertainty $u \in [0, 1]$.

% 对五个纠偏轴，我们按如下方式构造局部高斯先验 $p(\mathbf{z}_r) = \mathcal{N}(\boldsymbol{\mu}_0, \boldsymbol{\Sigma}_0)$：
For the five corrective axes, we construct a local Gaussian prior $p(\mathbf{z}_r) = \mathcal{N}(\boldsymbol{\mu}_0, \boldsymbol{\Sigma}_0)$ as follows:
\begin{IEEEeqnarray}{rCl}
  \mu_{0,k} &=& d_k \cdot s_k, \label{eq:prior_mean} \\
  \sigma_{0,k}^2 &=& \sigma_{\min}^2 + (\sigma_{\max}^2 - \sigma_{\min}^2) \cdot \frac{(1 - c_k) + u}{2}, \label{eq:prior_var} \\
  \boldsymbol{\Sigma}_0 &=& \mathrm{diag}(\sigma_{0,1}^2, \ldots, \sigma_{0,5}^2). \label{eq:prior_cov}
\end{IEEEeqnarray}
% 置信度越高且全局不确定性越低，先验方差越小，先验对后验的约束力越强；反之先验趋于弥散，后验更多依赖在线证据。
Higher confidence and lower global uncertainty yield smaller prior variance, giving the prior stronger influence on the posterior; otherwise the prior becomes diffuse and the posterior relies more on online evidence.

% 对 trust 轴，负方向会被截断为零，而 strength 与 direction 的乘积给出 $\tau_{\text{prior}} \in [0,1]$。一个学习得到的异常头（离线训练；见 \cref{sec:3-3}）还会根据首窗观测特征估计诊断异常分数 $\tau_{\text{diag}} \in [0,1]$。episode 级观测异常分数融合这两个来源：
For the trust axis, the negative direction is clipped to zero, and the product of strength and direction yields $\tau_{\text{prior}} \in [0,1]$. A learned anomaly head (trained offline; see \cref{sec:3-3}) also estimates a diagnostic anomaly score $\tau_{\text{diag}} \in [0,1]$ from the first-window observation features. The episode-level observation anomaly score fuses both sources:
\begin{IEEEeqnarray}{rCl}
  \tau_r &=& 1 - (1 - \tau_{\text{prior}})(1 - \tau_{\text{diag}}). \label{eq:trust}
\end{IEEEeqnarray}
% $\tau_r$ 在整个 episode 中保持不变，仅作用于观测侧校正的门控强度。
$\tau_r$ remains constant throughout the episode and only affects the gating strength of observation-side corrections.

\subsection{Offline Learning: Semantic-Anchored Adaptation Manifold}\label{sec:3-3}

% 离线阶段学习一组低容量线性映射，使五维 latent 的每个方向都解释一种可解释的工业失配类别。它还会学习诊断头以及训练分布的 support geometry。
The offline phase learns a set of low-capacity linear mappings so that each direction of the five-dimensional latent explains one interpretable class of industrial mismatch. It also learns diagnostic heads and the training-distribution support geometry.

% 训练数据来自两个来源。合成锚点对训练样本施加受控扰动，包括观测偏置、观测比例缩放、动力学加速/减速和观测时延。这些扰动会生成沿已知语义方向变化的样本，为 bias、scale 和 dynamics 轴提供几何骨架。
Training data comes from two sources. Synthetic anchors apply controlled perturbations to training samples, including observation bias, observation scaling, dynamics acceleration/deceleration, and observation delay. These perturbations produce perturbed samples along known semantic directions that provide the geometric skeleton for the bias, scale, and dynamics axes.

% Load 和 readout 轴的语义对齐主要由真实训练 episode 的残差与上下文结构提供，因为这两类失配难以通过简单的信号级扰动合成。真实训练 episode 残差序列从训练 episode 中提取主干预测的残差、观测特征、summary 特征和上下文特征。通过交替推断 corrective latent 与重拟合线性映射，适配流形同时贴合合成锚点的语义方向和真实数据的残差结构。
The semantic alignment of the load and readout axes comes mainly from the residual and context structure of real training episodes, because these two types of mismatch are difficult to synthesize via simple signal-level perturbations. Real training episode residual sequences provide backbone prediction residuals, observation features, summary features, and context features. By alternating corrective-latent inference with linear-mapping refitting, the adaptation manifold aligns with both the semantic directions of synthetic anchors and the residual structure of real data.

% 离线阶段产出一个 latent adapter model，其中包含以下可学习组件：
The offline phase produces a latent adapter model with the following learnable components:

% 轴系数映射 $\mathbf{b}_t = \mathbf{W} \tilde{\mathbf{h}}_t + \mathbf{a}$：将标准化的主干隐藏态 $\tilde{\mathbf{h}}_t$（使用训练集统计量做零均值单位方差标准化）线性投影为五维轴系数向量。
Axis coefficient mapping $\mathbf{b}_t = \mathbf{W} \tilde{\mathbf{h}}_t + \mathbf{a}$: a linear projection from the standardized backbone hidden state $\tilde{\mathbf{h}}_t$ (zero-mean, unit-variance normalization using training-set statistics) to a five-dimensional axis coefficient vector.

% 观测头：将观测异常特征（预测偏离、特征漂移、通道敏感性偏移等）映射到 bias/scale 子空间的 latent 观测值，同时估计标量异常分数 $\tau_{\text{diag}}$。
Observation heads: map observation anomaly features (prediction deviation, feature drift, channel sensitivity shift, etc.) to latent observations in the bias/scale subspace, and simultaneously estimate the scalar anomaly score $\tau_{\text{diag}}$.

% 标称 summary 预测：从隐藏态预测训练分布下的期望 summary 特征；在线阶段用实际 summary 与标称预测之间的残差作为 dynamics 证据。
Nominal summary prediction: predicts the expected summary features under the training distribution from hidden states; the online phase uses the residual between actual and nominal summaries as dynamics evidence.

% Load / Dynamics / Readout 子空间头：分别从对应的上下文或残差化特征估计各子空间的 latent 观测值。
Load / dynamics / readout subspace heads: estimate latent observations for each subspace from the corresponding context or residualized features.

% Support geometry：它由合成锚点 latent 与训练 episode 推断出的 corrective latent 共同形成一个 support latents 集合，并被标准化为近似零均值和单位对角协方差。在线阶段通过后验均值到 support latents 的最近距离来度量偏离训练分布的程度。
Support geometry: the set of support latents formed by synthetic anchor latents and inferred training episode corrective latents, standardized to approximately zero mean and unit diagonal covariance. The online phase measures deviation from the training distribution via the nearest distance from the posterior mean to the support latents.

% 以轴系数映射为例，训练目标为标准的岭回归：
For example, the axis coefficient mapping is fitted by ridge regression:
\begin{IEEEeqnarray}{l}
  \min_{\mathbf{W}, \mathbf{a}} \sum_{t} \| e_t - (\mathbf{W}\tilde{\mathbf{h}}_t + \mathbf{a})^\top \mathbf{z}_r \|^2 + \lambda \|\mathbf{W}\|_F^2, \label{eq:ridge}
\end{IEEEeqnarray}
% 其中 $e_t = y_t - \hat{y}_t^0$ 是主干残差，$\mathbf{z}_r$ 是当前 episode 的 corrective latent（对合成锚点而言是已知的，对真实 episode 而言由交替过程推断得到）。观测头、summary 预测头和各子空间头都采用类似的线性回归形式，但其输入特征和目标子空间各不相同。
where $e_t = y_t - \hat{y}_t^0$ is the backbone residual and $\mathbf{z}_r$ is the current episode's corrective latent (known for synthetic anchors, inferred by alternation for real episodes). The observation heads, summary prediction head, and subspace heads all adopt a similar linear regression form, but differ in their input features and target subspaces.

% 联合训练首先在合成锚点上拟合初始轴系数映射和诊断头。若提供训练 episode 数据，ROAM 会交替执行 corrective-latent 推断与映射重拟合，共进行 $N_{\text{iter}}$ 轮，直到映射残差稳定。随后，所有线性头会在合并后的锚点和训练 episode 上联合重拟合。所有映射都是线性的，并且所有离线学习到的组件在在线适配期间都完全冻结。
Joint training first fits the initial axis coefficient mapping and diagnostic heads on synthetic anchors. If training episode data are available, ROAM alternates corrective-latent inference with mapping refitting for $N_{\text{iter}}$ rounds until the mapping residual stabilizes. All linear heads are then jointly refitted on the combined set of anchors and training episodes. Every mapping is linear, and all offline-learned components are fully frozen during online adaptation.

\subsection{Online Posterior Update: Structured Evidence Fusion}\label{sec:3-4}

\begin{figure}[t]
  \centering
  \includegraphics[width=88mm]{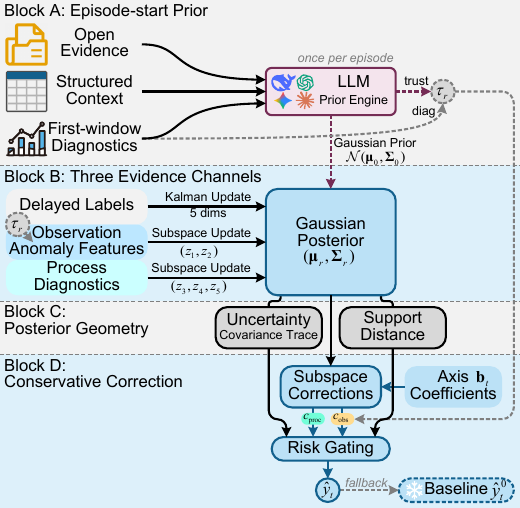}
  % 在线后验更新与保守校正。(A) LLM 先验引擎在每个 episode 处理一次开放式证据、结构化上下文和首窗诊断，生成高斯先验与观测异常分数 $\tau_r$。(B) 延迟标签、观测异常特征和过程诊断会在每个预测步都在各自的语义子空间内更新共享后验。(C) 提取后验不确定性与支撑距离用于风险控制。(D) 子空间校正通过多层风险门控得到适配预测；虚线箭头表示证据不足时回退到冻结基线。
  \caption{Online posterior update and conservative correction. (A)~The LLM prior engine processes open-ended evidence, structured context, and first-window diagnostics once per episode to produce a Gaussian prior and an observation anomaly score~$\tau_r$. (B)~Delayed labels, observation anomaly features, and process diagnostics each update the shared posterior within their respective semantic subspaces at each prediction step. (C)~Posterior uncertainty and support distance are extracted for risk control. (D)~Subspace corrections pass through multi-layer risk gating to yield the adapted prediction; dashed arrows indicate fallback to the frozen baseline when evidence is insufficient.}
  \label{fig:online_detail}
\end{figure}

% 在线阶段维护一个关于 $\mathbf{z}_r$ 的高斯后验 $\mathcal{N}(\boldsymbol{\mu}_r, \boldsymbol{\Sigma}_r)$，初始化为 \cref{sec:3-2} 中构造的先验 $(\boldsymbol{\mu}_0, \boldsymbol{\Sigma}_0)$。随 episode 推进，三类异质证据通过不同路径融入后验，每类证据只作用于对应的语义子空间。\cref{fig:online_detail} 展示了从先验构造到最终保守校正的完整数据流。
The online phase maintains a Gaussian posterior $\mathcal{N}(\boldsymbol{\mu}_r, \boldsymbol{\Sigma}_r)$ over $\mathbf{z}_r$, initialized from the prior $(\boldsymbol{\mu}_0, \boldsymbol{\Sigma}_0)$ constructed in \cref{sec:3-2}. As the episode progresses, three types of heterogeneous evidence are fused into the posterior through different pathways, with each type acting only on its corresponding semantic subspace. \cref{fig:online_detail} illustrates the complete data flow from prior construction to the final conservative correction.

% \textit{延迟标签残差。} 当历史标签在 $\Delta$ 步延迟后揭示时，我们计算残差 $e_t = y_t - \hat{y}_t^0$，以当前时刻的轴系数向量 $\mathbf{b}_t$ 为观测矩阵，做一次标准的线性-高斯 Kalman 单步更新：
\textit{Delayed label residuals.} When historical labels are revealed after a delay of $\Delta$ steps, we compute the residual $e_t = y_t - \hat{y}_t^0$ and perform a standard linear-Gaussian Kalman single-step update using the current axis coefficient vector $\mathbf{b}_t$ as the observation matrix:
\begin{IEEEeqnarray}{rCl}
  \mathbf{K}_t &=& \frac{\boldsymbol{\Sigma}_r \mathbf{b}_t}{\mathbf{b}_t^\top \boldsymbol{\Sigma}_r \mathbf{b}_t + \sigma_y^2}, \label{eq:kalman_gain} \\
  \boldsymbol{\mu}_r &\leftarrow& \boldsymbol{\mu}_r + \mathbf{K}_t (e_t - \mathbf{b}_t^\top \boldsymbol{\mu}_r), \label{eq:kalman_mean} \\
  \boldsymbol{\Sigma}_r &\leftarrow& \boldsymbol{\Sigma}_r - \mathbf{K}_t \mathbf{b}_t^\top \boldsymbol{\Sigma}_r, \label{eq:kalman_cov}
\end{IEEEeqnarray}
% 其中 $\mathbf{K}_t$ 为 Kalman 增益，$\sigma_y^2$ 为标签观测噪声方差。这是最直接的证据来源——真实标签直接驱动 latent 向解释残差的方向移动。
where $\mathbf{K}_t$ is the Kalman gain and $\sigma_y^2$ is the label observation noise variance. This is the most direct evidence source---real labels drive the latent toward the direction that explains the residual.

% \textit{子空间高斯更新。} 观测证据和过程证据都通过下面的一般形式融入后验；这是一种限制在索引集 $\mathcal{I}$ 上的类似高斯更新。设 $\tilde{\mathbf{z}}$ 为该子空间内的 latent 观测值，$\sigma^2$ 为观测噪声方差：
\textit{Subspace Gaussian update.} Both observation and process evidence are fused into the posterior via the following general form, an analogous Gaussian update restricted to index set $\mathcal{I}$. Let $\tilde{\mathbf{z}}$ denote the latent observation in that subspace and $\sigma^2$ the observation noise variance:
\begin{IEEEeqnarray}{rCl}
  \mathbf{K}_{\mathcal{I}} &=& \boldsymbol{\Sigma}_r[\mathcal{I},\mathcal{I}] \big(\boldsymbol{\Sigma}_r[\mathcal{I},\mathcal{I}] + \sigma^2 \mathbf{I}\big)^{-1}, \label{eq:sub_gain} \\
  \boldsymbol{\mu}_r[\mathcal{I}] &\leftarrow& \boldsymbol{\mu}_r[\mathcal{I}] + \mathbf{K}_{\mathcal{I}} \big(\tilde{\mathbf{z}} - \boldsymbol{\mu}_r[\mathcal{I}]\big), \label{eq:sub_mean} \\
  \boldsymbol{\Sigma}_r[\mathcal{I},\mathcal{I}] &\leftarrow& (\mathbf{I} - \mathbf{K}_{\mathcal{I}}) \boldsymbol{\Sigma}_r[\mathcal{I},\mathcal{I}]. \label{eq:sub_cov}
\end{IEEEeqnarray}
% 该更新仅修改 $\boldsymbol{\mu}_r$ 和 $\boldsymbol{\Sigma}_r$ 在 $\mathcal{I}$ 上的分量，其余维度不受影响。
This update modifies only the components of $\boldsymbol{\mu}_r$ and $\boldsymbol{\Sigma}_r$ at indices $\mathcal{I}$; all other dimensions remain unchanged.

% \textit{观测异常证据。} 观测异常特征（特征漂移、通道敏感性偏移等）经学习的观测头映射为 bias/scale 子空间中的 latent 观测值 $\tilde{\mathbf{z}}_{\text{obs}}$。在预定义的更新间隔内对缓冲区特征求平均后，以 $\mathcal{I}_{\text{obs}} = \{1, 2\}$ 执行上述子空间更新。观测证据的有效噪声方差会被 $\tau_r$ 放大：
\textit{Observation anomaly evidence.} Observation anomaly features (feature drift, channel sensitivity shift, etc.) are mapped by the learned observation head to a latent observation $\tilde{\mathbf{z}}_{\text{obs}}$ in the bias/scale subspace. After averaging buffered features over a predefined update interval, the subspace update above is applied with $\mathcal{I}_{\text{obs}} = \{1, 2\}$. The effective noise variance of observation evidence is amplified by $\tau_r$:
\begin{IEEEeqnarray}{rCl}
  \sigma_{\text{obs,eff}}^2 &=& \sigma_{\text{obs}}^2 (1 + \tau_r), \label{eq:obs_noise}
\end{IEEEeqnarray}
% 使得观测异常程度越高时，观测证据对后验的拉动越弱。
so that higher observation anomaly weakens the pull of observation evidence on the posterior.

% \textit{过程诊断证据。} 过程相关证据分为三条独立的子空间通道，各自使用上述子空间更新（\cref{eq:sub_gain}--\cref{eq:sub_cov}）。(i) Dynamics 证据来自当前窗口的 summary 特征与标称 summary 预测之间的残差，经 dynamics 头映射后以噪声方差 $\sigma_{\text{dyn}}^2$ 更新 $z_4$。(ii) Readout 证据来自残差与上下文特征的交互项，经 readout 头映射后以噪声方差 $\sigma_{\text{read}}^2$ 更新 $z_5$。(iii) Load 证据来自上下文特征结合当前残差均值，经 load 头映射后以噪声方差 $\sigma_{\text{load}}^2$ 更新 $z_3$。与观测证据相同，过程证据也按预定义间隔做缓冲聚合后批量更新。
\textit{Process diagnostic evidence.} Process-related evidence is split into three independent subspace channels, each using the subspace update in \cref{eq:sub_gain}--\cref{eq:sub_cov}. (i)~Dynamics evidence comes from the residual between the current window's summary features and the nominal summary prediction; the dynamics head maps it to update $z_4$ with noise variance $\sigma_{\text{dyn}}^2$. (ii)~Readout evidence comes from the interaction between residuals and context features; the readout head maps it to update $z_5$ with noise variance $\sigma_{\text{read}}^2$. (iii)~Load evidence comes from context features combined with the current residual mean; the load head maps it to update $z_3$ with noise variance $\sigma_{\text{load}}^2$. Like observation evidence, process evidence is buffered and batch-updated at predefined intervals.

% 在每个预测时刻，从后验导出两类几何量供风险控制使用：后验协方差迹（posterior trace），分别在观测子空间和过程子空间上度量不确定性大小；以及 support distance，即后验均值到最近 support latent 的欧氏距离，度量偏离训练分布的程度。
At each prediction step, two types of geometric quantities are derived from the posterior for risk control: the posterior trace, measuring uncertainty magnitude separately on the observation and process subspaces; and the support distance, the Euclidean distance from the posterior mean to the nearest support latent, measuring deviation from the training distribution.

\subsection{Output Correction and Risk Gating}\label{sec:3-5}

% 给定后验 $(\boldsymbol{\mu}_r, \boldsymbol{\Sigma}_r)$ 和当前时刻的轴系数 $\mathbf{b}_t$，ROAM 通过子空间点积计算校正量，并通过多层门控将其加到基线预测上（对应 \cref{fig:online_detail} 中的 Block D）。设基线预测为 $\hat{y}_t^0$。轴系数映射给出 $\mathbf{b}_t \in \mathbb{R}^5$，其被划分为观测部分 $\mathbf{b}_t^{\text{obs}}$（bias/scale 轴）和过程部分 $\mathbf{b}_t^{\text{proc}}$（load/dynamics/readout 轴）。子空间校正量为：
Given the posterior $(\boldsymbol{\mu}_r, \boldsymbol{\Sigma}_r)$ and the current axis coefficients $\mathbf{b}_t$, ROAM computes corrections via subspace dot products and applies them to the baseline prediction through multi-layer gating (Block~D in \cref{fig:online_detail}). Let the baseline prediction be $\hat{y}_t^0$. The axis coefficient mapping yields $\mathbf{b}_t \in \mathbb{R}^5$, split into an observation part $\mathbf{b}_t^{\text{obs}}$ (bias/scale axes) and a process part $\mathbf{b}_t^{\text{proc}}$ (load/dynamics/readout axes). The subspace corrections are:
\begin{IEEEeqnarray}{rCl}
  c_{\text{obs}} &=& (\mathbf{b}_t^{\text{obs}})^\top \boldsymbol{\mu}_r^{\text{obs}}, \label{eq:c_obs} \\
  c_{\text{proc}} &=& (\mathbf{b}_t^{\text{proc}})^\top \boldsymbol{\mu}_r^{\text{proc}}, \label{eq:c_proc}
\end{IEEEeqnarray}
% 其中 $\boldsymbol{\mu}_r^{\text{obs}}$ 和 $\boldsymbol{\mu}_r^{\text{proc}}$ 为后验均值在对应子空间的投影。
where $\boldsymbol{\mu}_r^{\text{obs}}$ and $\boldsymbol{\mu}_r^{\text{proc}}$ are the posterior mean projected onto the corresponding subspaces.

% 每个子空间还有一个效果强度系数。当后验均值相对于不确定性很小时，它会趋近于零；即使校正方向是正确的，置信度不足也不会产生大幅校正：
Each subspace has an effect strength coefficient that approaches zero when the posterior mean is small relative to the uncertainty---even if the correction direction is correct, insufficient confidence prevents large corrections:
\begin{IEEEeqnarray}{rCl}
  \alpha_{\text{obs}} &=& \frac{\|\boldsymbol{\mu}_r^{\text{obs}}\|}{\|\boldsymbol{\mu}_r^{\text{obs}}\| + \sqrt{\mathrm{tr}(\boldsymbol{\Sigma}_r^{\text{obs}})} + \epsilon}, \label{eq:alpha_obs}
\end{IEEEeqnarray}
% $\alpha_{\text{proc}}$ 同理，$\epsilon > 0$ 为数值稳定性常数。
with $\alpha_{\text{proc}}$ defined analogously and $\epsilon > 0$ a small constant for numerical stability.

% 基础门控系数 $\gamma$ 由后验协方差迹和 support distance 联合决定：
The base gating coefficient $\gamma$ is jointly determined by the posterior covariance trace and the support distance:
\begin{IEEEeqnarray}{l}
  \gamma = \exp\!\Big({-}\eta_1 \cdot \mathrm{tr}(\boldsymbol{\Sigma}_r^{\text{obs}}) - \eta_2 \cdot \mathrm{tr}(\boldsymbol{\Sigma}_r^{\text{proc}}) \nonumber \\
  \qquad\qquad {-}\beta_1 \cdot d_{\text{obs}} - \beta_2 \cdot d_{\text{proc}}\Big), \label{eq:gamma}
\end{IEEEeqnarray}
% 其中 $d_{\text{obs}}$ 和 $d_{\text{proc}}$ 分别表示在观测子空间和过程子空间内，后验均值到最近 support latent 的欧氏距离（度量偏离训练分布的程度），$\eta_1, \eta_2, \beta_1, \beta_2$ 为超参数。观测侧和过程侧的门控系数分别为：
where $d_{\text{obs}}$ and $d_{\text{proc}}$ denote the nearest support-latent distances computed in the observation and process subspaces, respectively (measuring deviation from the training distribution), and $\eta_1, \eta_2, \beta_1, \beta_2$ are hyperparameters. The observation-side and process-side gating coefficients are:
\begin{IEEEeqnarray}{rCl}
  \gamma_{\text{obs}} &=& \frac{\gamma}{1 + \tau_r}, \qquad \gamma_{\text{proc}} = \gamma. \label{eq:gamma_split}
\end{IEEEeqnarray}
% 观测异常分数 $\tau_r$ 只抑制观测相关校正。最终适配预测为：
The observation anomaly score $\tau_r$ suppresses only observation-related corrections. The final adapted prediction is:
\begin{IEEEeqnarray}{rCl}
  \hat{y}_t &=& \hat{y}_t^0 + \gamma_{\text{obs}} \cdot \alpha_{\text{obs}} \cdot c_{\text{obs}} + \gamma_{\text{proc}} \cdot \alpha_{\text{proc}} \cdot c_{\text{proc}}. \label{eq:adapted_pred}
\end{IEEEeqnarray}

% 先验扩散、效果强度抑制、门控衰减以及 trust 门控会层层级联，使系统在不确定情境下趋向 no-op，即退化为原始主干预测。所有先验判断、后验更新和门控系数都可以被记录下来，用于事后审查。
Prior diffusion, effect strength suppression, gating decay, and trust gating cascade to drive the system toward no-op under uncertainty, i.e., it degrades to the original backbone prediction. All prior judgments, posterior updates, and gating coefficients can be logged for post-hoc inspection.

\section{Experiments}\label{sec:4}

% 我们在两个工业过程场景上评估 ROAM。
We evaluate ROAM on two industrial processes.

\subsection{Experimental Settings}\label{sec:4-1}

% 所有主干和 ROAM 组件都在同一代码库中以相同的超参数训练和评估。不对不同数据集或不同主干进行单独调参。
All backbones and ROAM components are trained and evaluated in the same codebase with identical hyperparameters. No per-dataset or per-backbone tuning is applied.
% 实验在一台配备 Intel Xeon 8470Q CPU、90\,GB RAM 和单张 NVIDIA RTX 5090 GPU（32\,GB）的工作站上运行。实现使用 Python~3.14、PyTorch~2.11.0 和 CUDA~13.0。主结果报告三次独立运行的平均值。
Experiments run on a workstation equipped with an Intel Xeon 8470Q CPU, 90\,GB RAM, and a single NVIDIA RTX 5090 GPU (32\,GB). The implementation uses Python~3.14, PyTorch~2.11.0, and CUDA~13.0. Main results report the mean over three independent runs.
% 每个主干都使用 Adam（lr\,=\,$1{\times}10^{-3}$）在固定 300 个 epoch 的预算下训练，然后冻结。ROAM 离线适配器通过 ridge regression（$\lambda{=}0.001$）在冻结主干上拟合，并在合成锚点和真实 episode 上交替进行 3 次迭代。在线先验引擎使用 DeepSeek-V3.2-Thinking。推理时不使用后处理。
Each backbone is trained with Adam (lr\,=\,$1{\times}10^{-3}$) for a fixed budget of 300 epochs and then frozen. The ROAM offline adapter is fitted on the frozen backbone via ridge regression ($\lambda{=}0.001$) with 3 alternating iterations over synthetic anchors and real episodes. The online prior engine uses DeepSeek-V3.2-Thinking. No post-processing is applied at inference.

% 我们报告三个指标：平均绝对误差（MAE~$\downarrow$）、均方根误差（RMSE~$\downarrow$）和决定系数（$R^2$~$\uparrow$）。效率用每步推理时间（ms/step~$\downarrow$）衡量。
We report three metrics: mean absolute error (MAE~$\downarrow$), root mean squared error (RMSE~$\downarrow$), and coefficient of determination ($R^2$~$\uparrow$). Efficiency is measured by per-step inference time (ms/step~$\downarrow$).
% 为了验证主干无关的设计，我们测试了七种主干，覆盖四类架构：经典机器学习（SVR、XGBoost）、循环网络（GRU、LSTM）、基于注意力的模型（Transformer、Informer）和状态空间模型（Mamba）。所有主干共享相同的 ROAM 适配器配置和评估协议。
To validate the backbone-agnostic design, we test seven backbones spanning four architecture families: classic ML (SVR~\cite{smola2004svr} and XGBoost~\cite{chen2016xgboost}), recurrent models (GRU~\cite{cho2014gru} and LSTM~\cite{hochreiter1997lstm}), attention-based models (Transformer~\cite{vaswani2017transformer} and Informer~\cite{zhou2021informer}), and a state-space model (Mamba~\cite{gu2024mamba}). All backbones share the same ROAM adapter configuration and evaluation protocol.

\subsection{Datasets}\label{sec:4-2}

% 我们在两个工业过程上验证 ROAM。在两个环境中，训练集都只包含标准生产运行，因此模型只学习标称过程动态，而不接触任何偏移或故障工况。这种设计更严格地检验了适配能力。每个测试集都保留一个分布内场景作为基线对照。\cref{tab:scenario_taxonomy} 给出了完整的场景分类。
We validate ROAM on two industrial processes. In both environments, the training set contains only standard production runs, so that the model learns exclusively nominal process dynamics without exposure to any shift or fault condition. This design provides a stricter test of adaptation capability. Each test set reserves one in-distribution scenario as a baseline control. \cref{tab:scenario_taxonomy} gives the full taxonomy.

\begin{table}[t]
  \centering
  % 浓密环境与 IndPenSim 两个环境的场景分类，按训练/测试划分并按偏移类别分组。
  \caption{Scenario taxonomy for the thickening and IndPenSim environments, organized by train/test split and shift category.}
  \label{tab:scenario_taxonomy}
  \setlength{\tabcolsep}{4pt}\scriptsize
  \begin{tabularx}{\linewidth}{l l l}
    \toprule
    Usage & Category                              & Scenario Description                                       \\
    \midrule
    \multicolumn{3}{l}{\textit{Thickening Dewatering (19 scenarios)}}                                          \\
    Train & Nominal                               & Feed jitter variations under nominal conditions $\times$ 6 \\
    \multirow[t]{14}{*}{Test}
          & Same-family                           & Held-out nominal feed-jitter                               \\
          & \multirow[t]{2}{*}{Visible shift}     & Sustained high upstream load                               \\
          &                                       & Sustained low upstream load                                \\
          & \multirow[t]{7}{*}{Hidden shift}      & Discharge efficiency drop                                  \\
          &                                       & Slurry property shift                                      \\
          &                                       & Feed solids ratio shift                                    \\
          &                                       & Flocculant dosing failure                                  \\
          &                                       & Media permeability loss                                    \\
          &                                       & Actuator deadtime / gain loss                              \\
          &                                       & Restart inventory shift                                    \\
          & \multirow[t]{3}{*}{Observation shift} & Pressure sensor bias                                       \\
          &                                       & Flow scale error                                           \\
          &                                       & Pressure signal delay                                      \\
    \midrule
    \multicolumn{3}{l}{\textit{IndPenSim Penicillin Fermentation (44 scenarios)}}                              \\
    Train & Recipe control                        & Standard recipe-based control $\times$ 29                  \\
    \multirow[t]{7}{*}{Test}
          & Same-family                           & Held-out standard-recipe                                   \\
          & Operator control                      & Operator-controlled process $\times$ 2                     \\
          & Control-regime                        & Raman-assisted advanced process control $\times$ 2         \\
          & \multirow[t]{4}{*}{Process fault}
          & Substrate feed anomaly $\times$ 3                                                                  \\
          &                                       & pH measurement drift $\times$ 2                            \\
          &                                       & Cooling / thermal anomaly $\times$ 2                       \\
          &                                       & Aeration anomaly $\times$ 3                                \\
    \bottomrule
  \end{tabularx}
\end{table}

\begin{figure*}[t]
  \centering
  \includegraphics[width=172mm]{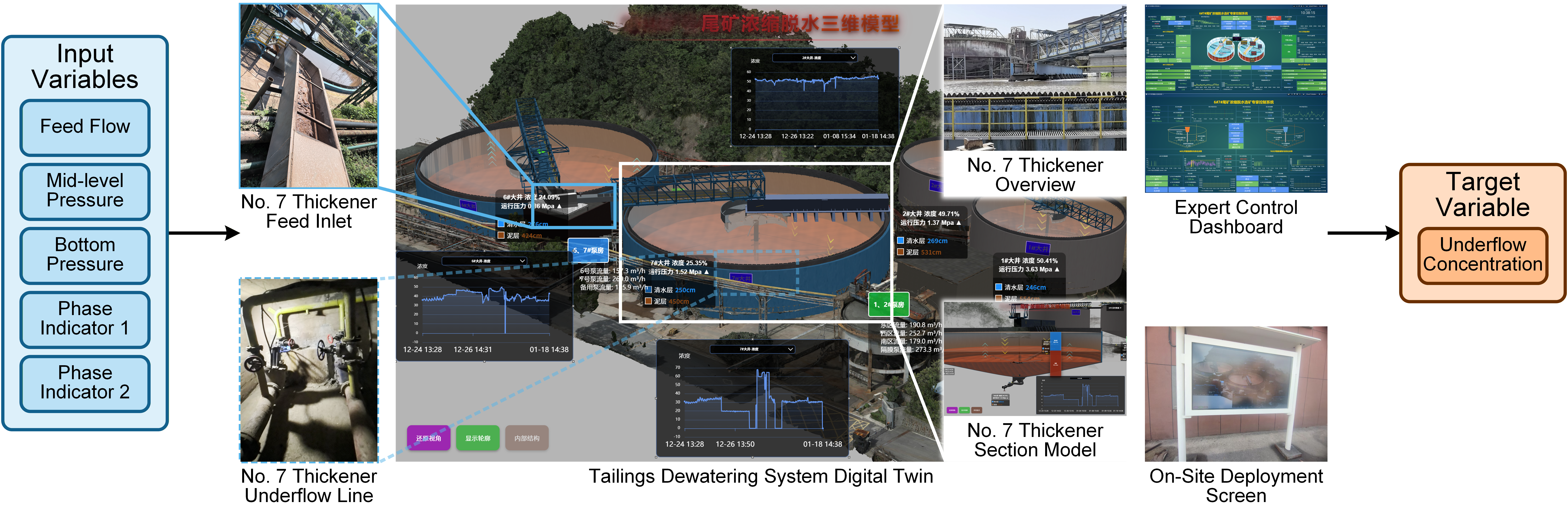}
  % 中国南京某厂 7\# 尾矿浓密脱水现场，含系统输入信号、数字孪生界面与现场部署视图。
  \caption{No.\,7 tailings thickening dewatering site at a plant in Nanjing, China, showing system input signals, the digital-twin interface, and on-site deployment views.}
  \label{fig:thickening_site}
\end{figure*}

% 浓密脱水环境来自中国南京某厂的 7\,\# 尾矿浓密机（直径 50\,m）。软测量目标是分钟级底流浓度 $c_{\text{uf}}$。该浓密机通过重力沉降和絮凝对尾矿浆进行脱水，而底流浓度直接决定下游压滤效率和尾矿库安全。模型输入包括入料流量、中层和底部压力以及两个相位指示量（共 5 个变量），每个 episode 持续 300~分钟。六个训练场景覆盖标称来料抖动，十四个测试场景覆盖一个基线对照、一个 same-family 泛化检查以及按可见性分组的偏移场景。\cref{fig:thickening_site} 展示了系统输入、数字孪生界面和现场部署视图。
The thickening dewatering environment originates from a No.\,7 tailings thickener (diameter 50\,m) at a plant in Nanjing, China. The soft-sensing target is minute-level underflow concentration $c_{\text{uf}}$. The thickener dewaters tailings slurry via gravity settling and flocculation, and underflow concentration directly governs downstream filter-press efficiency and tailings storage safety. Model inputs are feed flow, mid-level and bottom pressures, and two phase indicators (5 variables), with each episode spanning 300~minutes. Six training scenarios cover nominal feed jitter, and fourteen test scenarios span a baseline control, a same-family generalization check, and shift scenarios grouped by visibility. \cref{fig:thickening_site} shows the system inputs, the digital-twin interface, and on-site deployment views.

% IndPenSim 环境来自公开可用的 IndPenSim V3 数据集~\cite{goldrick2015development}，该数据集模拟了工业规模的青霉素补料分批发酵过程。该过程在生物反应器中通过受控的温度、pH、溶解氧和补料策略培养 \textit{Penicillium chrysogenum}。软测量目标是离线青霉素浓度。模型输入由 23 个在线过程变量组成，窗口长度为 120 步。二十九个训练批次覆盖标准配方式控制。十五个测试场景覆盖一个基线对照、操作员控制运行、拉曼辅助 APC 运行以及过程故障。
The IndPenSim environment is drawn from the publicly available IndPenSim V3 dataset~\cite{goldrick2015development}, which simulates an industrial-scale penicillin fed-batch fermentation process. The process cultivates \textit{Penicillium chrysogenum} in a bioreactor under controlled temperature, pH, dissolved oxygen, and feeding strategies. The soft-sensing target is offline penicillin concentration. Model inputs comprise 23 online process variables with a window length of 120 steps. Twenty-nine training batches cover standard recipe-based control. Fifteen test scenarios span a baseline control, operator-controlled runs, Raman-assisted APC runs, and process faults.

\subsection{Main Results}\label{sec:4-3}

% \cref{tab:main_thickening} 和 \cref{tab:main_indpensim} 报告了每种方法在各场景类别上的 MAE、RMSE、$R^2$ 和每步推理时间。
\cref{tab:main_thickening} and \cref{tab:main_indpensim} report MAE, RMSE, $R^2$, and per-step inference time for each method across scenario categories.

\begin{table*}[t]
  \centering
  % 浓密环境上的定量对比（七种主干 $\times$ Base/+ROAM）。MAE 与 RMSE 以 $\times 10^{-3}$ 报告。各列最优值加粗。
  \caption{Quantitative comparison in the thickening environment across seven backbones. MAE and RMSE are in units of \texttimes\,10\textsuperscript{3}. Best per column in bold.}
  \label{tab:main_thickening}
  \setlength{\tabcolsep}{4pt}\scriptsize
  \begin{tabularx}{\linewidth}{ll *{12}{r} c}
    \toprule
                                 &         & \multicolumn{3}{c}{Same-family} & \multicolumn{3}{c}{Visible shift} & \multicolumn{3}{c}{Hidden shift} & \multicolumn{3}{c}{Observation shift} &                                                                                                                                                                                            \\
    \cmidrule(lr){3-5}\cmidrule(lr){6-8}\cmidrule(lr){9-11}\cmidrule(lr){12-14}
    Backbone                     & Setting & MAE$\downarrow$                 & RMSE$\downarrow$                  & $R^2$$\uparrow$                  & MAE$\downarrow$                       & RMSE$\downarrow$ & $R^2$$\uparrow$  & MAE$\downarrow$ & RMSE$\downarrow$ & $R^2$$\uparrow$  & MAE$\downarrow$ & RMSE$\downarrow$ & $R^2$$\uparrow$  & Inference Time (ms/step)$\downarrow$ \\
    \midrule
    \multirow{2}{*}{SVR}         & Base    & 0.188                           & 0.231                             & 0.99969                          & 11.183                                & 12.963           & -0.03685         & 24.219          & 41.572           & 0.12918          & 5.255           & 7.685            & 0.66267          & \textbf{0.144}                       \\
                                 & +ROAM   & 0.185                           & 0.227                             & 0.99970                          & 8.518                                 & 10.002           & 0.38279          & 19.684          & 37.019           & 0.30949          & 4.122           & 6.013            & 0.79349          & 0.151                                \\
    \multirow{2}{*}{XGBoost}     & Base    & 0.047                           & 0.061                             & 0.99998                          & \textbf{0.059}                        & \textbf{0.079}   & \textbf{0.99996} & 23.128          & 40.630           & 0.16822          & \textbf{0.051}  & \textbf{0.065}   & \textbf{0.99998} & 0.255                                \\
                                 & +ROAM   & 0.047                           & 0.061                             & 0.99998                          & 0.060                                 & 0.080            & \textbf{0.99996} & 18.888          & 36.120           & 0.34263          & 0.059           & 0.078            & 0.99997          & 0.270                                \\
    \multirow{2}{*}{GRU}         & Base    & 0.027                           & 0.034                             & 0.99999                          & 0.672                                 & 0.846            & 0.99559          & 22.957          & 40.391           & 0.17796          & 2.341           & 3.482            & 0.93073          & 1.532                                \\
                                 & +ROAM   & 0.023                           & 0.031                             & 0.99999                          & 0.551                                 & 0.754            & 0.99649          & 18.329          & 35.154           & 0.37732          & 1.752           & 2.752            & 0.95674          & 1.538                                \\
    \multirow{2}{*}{LSTM}        & Base    & 0.036                           & 0.058                             & 0.99998                          & 0.968                                 & 1.300            & 0.98957          & 23.014          & 40.443           & 0.17584          & 2.907           & 4.392            & 0.88984          & 0.584                                \\
                                 & +ROAM   & 0.037                           & 0.057                             & 0.99998                          & 0.754                                 & 1.057            & 0.99311          & 18.282          & 35.021           & 0.38202          & 2.141           & 3.234            & 0.94027          & 0.589                                \\
    \multirow{2}{*}{Transformer} & Base    & 0.057                           & 0.082                             & 0.99996                          & 3.890                                 & 4.574            & 0.87090          & 22.977          & 40.424           & 0.17660          & 2.502           & 3.553            & 0.92789          & 1.170                                \\
                                 & +ROAM   & 0.055                           & 0.078                             & 0.99997                          & 3.668                                 & 4.371            & 0.88214          & 18.298          & 35.115           & 0.37868          & 2.665           & 4.568            & 0.88078          & 1.186                                \\
    \multirow{2}{*}{Informer}    & Base    & 0.400                           & 0.493                             & 0.99861                          & 0.837                                 & 0.964            & 0.99427          & 24.398          & 41.484           & 0.13288          & 1.824           & 3.514            & 0.92945          & 7.845                                \\
                                 & +ROAM   & 0.382                           & 0.470                             & 0.99874                          & 0.711                                 & 0.819            & 0.99587          & 19.382          & 36.079           & 0.34411          & 1.430           & 2.687            & 0.95875          & 7.858                                \\
    \multirow{2}{*}{Mamba}       & Base    & 0.019                           & \textbf{0.023}                    & \textbf{1.00000}                 & 7.316                                 & 8.475            & 0.55684          & 22.842          & 40.303           & 0.18154          & 3.400           & 4.192            & 0.89964          & 8.716                                \\
                                 & +ROAM   & \textbf{0.018}                  & \textbf{0.023}                             & \textbf{1.00000}                 & 5.369                                 & 6.379            & 0.74890          & \textbf{18.150} & \textbf{34.891}  & \textbf{0.38658} & 2.725           & 3.322            & 0.93698          & 8.736                                \\
    \bottomrule
  \end{tabularx}
\end{table*}

\begin{table*}[t]
  \centering
  % IndPenSim 环境上的定量对比（七种主干 $\times$ Base/+ROAM）。各列最优值加粗。
  \caption{Quantitative comparison on the IndPenSim environment across seven backbones. Best per column in bold.}
  \label{tab:main_indpensim}
  \setlength{\tabcolsep}{4pt}\scriptsize
  \begin{tabularx}{\linewidth}{ll *{12}{r} c}
    \toprule
                                 &         & \multicolumn{3}{c}{Same-family} & \multicolumn{3}{c}{Operator control} & \multicolumn{3}{c}{Control-regime} & \multicolumn{3}{c}{Process fault} &                                                                                                                                                                                            \\
    \cmidrule(lr){3-5}\cmidrule(lr){6-8}\cmidrule(lr){9-11}\cmidrule(lr){12-14}
    Backbone                     & Setting & MAE$\downarrow$                 & RMSE$\downarrow$                     & $R^2$$\uparrow$                    & MAE$\downarrow$                   & RMSE$\downarrow$ & $R^2$$\uparrow$  & MAE$\downarrow$ & RMSE$\downarrow$ & $R^2$$\uparrow$  & MAE$\downarrow$ & RMSE$\downarrow$ & $R^2$$\uparrow$  & Inference Time (ms/step)$\downarrow$ \\
    \midrule
    \multirow{2}{*}{SVR}         & Base    & 1.675                           & 2.158                                & 0.81036                            & 2.175                             & 3.994            & 0.82197          & 1.727           & 2.362            & 0.94252          & 3.360           & 4.431            & 0.72375          & 0.638                                \\
                                 & +ROAM   & 1.596                           & 2.042                                & 0.83028                            & 1.797                             & 2.998            & 0.89970          & 1.516           & 2.107            & 0.95424          & 2.775           & 3.665            & 0.81101          & 0.650                                \\
    \multirow{2}{*}{XGBoost}     & Base    & 1.345                           & 1.960                                & 0.84364                            & 0.860                             & 1.466            & 0.97601          & 1.127           & 1.635            & 0.97245          & 3.212           & 4.521            & 0.71246          & \textbf{0.283}                       \\
                                 & +ROAM   & 1.159                           & 1.750                                & 0.87536                            & \textbf{0.847}                    & 1.450            & 0.97653          & \textbf{1.050}  & 1.452            & 0.97826          & 2.420           & 3.320            & 0.84494          & 0.301                                \\
    \multirow{2}{*}{GRU}         & Base    & 0.928                           & 1.227                                & 0.93875                            & 0.925                             & 1.197            & 0.98400          & 1.300           & 1.828            & 0.96558          & 2.965           & 4.267            & 0.74380          & 0.529                                \\
                                 & +ROAM   & 0.938                           & 1.232                                & 0.93818                            & 0.897                             & \textbf{1.171}   & \textbf{0.98469} & 1.229           & 1.681            & 0.97089          & 2.640           & 3.736            & 0.80365          & 0.534                                \\
    \multirow{2}{*}{LSTM}        & Base    & 0.865                           & 1.160                                & 0.94520                            & 2.442                             & 4.107            & 0.81175          & 1.470           & 1.876            & 0.96373          & 3.070           & 4.565            & 0.70676          & 0.565                                \\
                                 & +ROAM   & \textbf{0.804}                  & 1.082                                & 0.95238                            & 1.791                             & 2.868            & 0.90822          & 1.346           & 1.732            & 0.96907          & 2.843           & 4.084            & 0.76533          & 0.569                                \\
    \multirow{2}{*}{Transformer} & Base    & 1.034                           & 1.313                                & 0.92979                            & 2.201                             & 3.479            & 0.86492          & 3.079           & 5.207            & 0.72054          & 2.696           & 3.742            & 0.80302          & 1.115                                \\
                                 & +ROAM   & 0.992                           & 1.262                                & 0.93517                            & 1.908                             & 2.874            & 0.90780          & 2.511           & 4.336            & 0.80621          & 2.364           & 3.295            & 0.84728          & 1.129                                \\
    \multirow{2}{*}{Informer}    & Base    & 0.815                           & 1.053                                & 0.95487                            & 2.355                             & 3.689            & 0.84808          & 1.086           & \textbf{1.388}   & \textbf{0.98015} & 2.066           & 3.166            & 0.85900          & 7.387                                \\
                                 & +ROAM   & 0.806                           & \textbf{1.010}                       & \textbf{0.95843}                   & 1.996                             & 3.059            & 0.89553          & 1.130           & 1.421            & 0.97918          & \textbf{1.821}  & \textbf{2.583}   & \textbf{0.90612} & 7.399                                \\
    \multirow{2}{*}{Mamba}       & Base    & 1.630                           & 2.194                                & 0.80403                            & 3.030                             & 4.348            & 0.78897          & 3.023           & 5.583            & 0.67876          & 3.801           & 5.110            & 0.63258          & 26.615                               \\
                                 & +ROAM   & 1.490                           & 2.085                                & 0.82306                            & 2.632                             & 3.560            & 0.85856          & 2.688           & 4.492            & 0.79208          & 3.371           & 4.668            & 0.69342          & 26.655                               \\
    \bottomrule
  \end{tabularx}
\end{table*}

% 在浓密环境上，所有主干的 same-family 性能都接近饱和，而 ROAM 带来的变化可以忽略。增益集中在 hidden shift 上。七个主干的 Base hidden-shift MAE 都落在狭窄的 22.8--24.4 区间内，而加入 ROAM 后一致降到 18.2--19.7，降幅约为 20\%，其中 Mamba+ROAM 取得最低值。Base 分数的紧密聚集表明，hidden shift 是一个与主干无关的瓶颈：load、dynamics 或 readout 的变化在原始通道中几乎不可见，因此所有主干都会退化到相近水平。
In the thickening environment, same-family performance is near saturation for all backbones, and ROAM introduces negligible change. The gains concentrate on hidden shift. Base hidden-shift MAE falls within a narrow 22.8--24.4 band across all seven backbones, dropping consistently to 18.2--19.7 with ROAM, a reduction of roughly 20\%, with Mamba+ROAM reaching the lowest value. The tight clustering of Base scores shows that hidden shift is a backbone-agnostic bottleneck: changes in load, dynamics, or readout are barely visible in raw channels, so all backbones degrade to a similar level.

% 相比之下，不同主干在 visible-shift 和 observation-shift 上的 Base MAE 会相差几个数量级。由于偏移信号已经编码在输入特征中，XGBoost Base 在这两类上的 MAE 已经低于其他所有主干的 +ROAM。observation-shift 的改进弱于 hidden-shift，而且一致性更差。部分污染会被主干自身的鲁棒性吸收，而 ROAM 只会在后验携带足够证据时介入。
Visible-shift and observation-shift Base MAEs, by contrast, vary by orders of magnitude across backbones. XGBoost Base already achieves lower MAE than every other backbone's +ROAM on both categories, because the shift signal is encoded in the input features. Observation-shift improvements are weaker and less uniform than hidden-shift gains. Some corruptions are absorbed by backbone robustness, and ROAM intervenes only where the posterior carries sufficient evidence.

% IndPenSim 在更困难的条件下验证了这一模式：绝对误差更大，主干排名也更不稳定。最明显的增益出现在 operator-control 和 process-fault 场景中，这两类场景的重训练数据最稀缺。control-regime 一列结果较为混合；一些 regime 变化已经编码在轨迹中，仅靠序列模型就能处理。推理时间给出了工程约束：树模型和 GRU/LSTM 的额外开销仍然实用，而 Informer 和 Mamba 的成本明显更高。在两个数据集上，ROAM 的收益都出现在失配被隐藏且重训练不现实的场景中。附加开销会随主干自身的计算开销增长。
IndPenSim confirms the pattern under harder conditions: absolute errors are larger and backbone rankings are less stable. The clearest gains appear in operator-control and process-fault scenarios, where retraining data are scarcest. The control-regime column is mixed; some regime changes are already encoded in trajectories and handled by sequence models alone. Inference time sets an engineering bound: tree models and GRU/LSTM keep overhead practical, while Informer and Mamba cost substantially more. Across both datasets, ROAM delivers its gains where mismatch is hidden and retraining is impractical. The added cost scales with the backbone's own computational footprint.

\subsection{Qualitative Results}\label{sec:4-4}

\begin{figure*}[t]
  \centering
  \includegraphics[width=\linewidth]{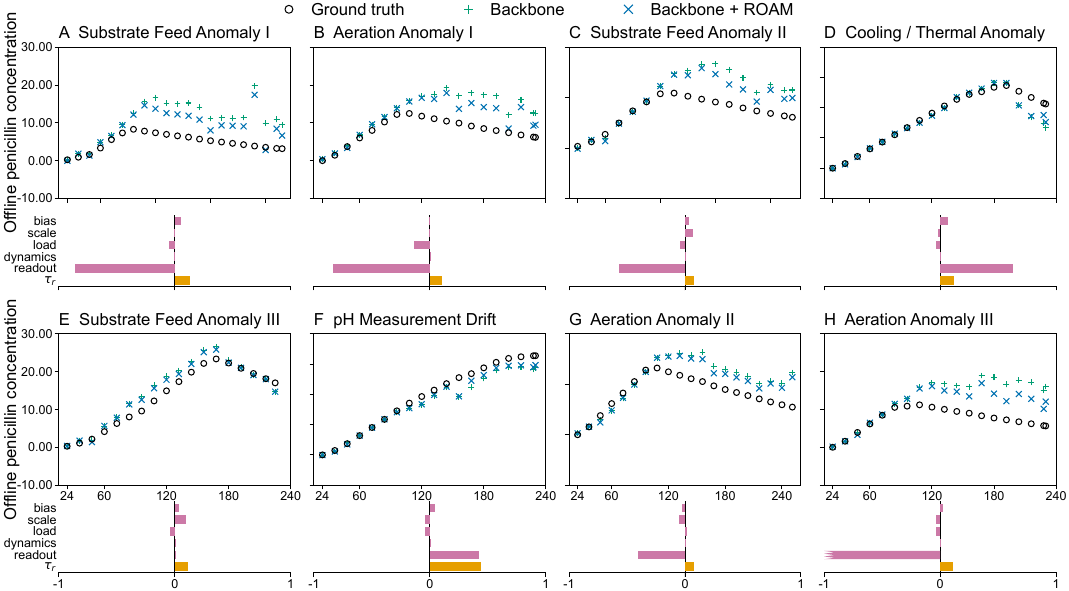}
  % 八个代表性的 IndPenSim 异常场景上的预测曲线（GRU Base vs. GRU+ROAM）。下方条带为五个语义轴与信任分数 $\tau_r$ 的最终后验值，锯齿端标记超出显示范围。
  \caption{Prediction curves on eight representative IndPenSim anomaly scenarios comparing GRU (Base) and GRU+ROAM. Lower strips report the final posterior on five semantic axes and the trust score $\tau_r$; serrated ends mark out-of-range values.}
  \label{fig:prediction_curves}
\end{figure*}

% \cref{fig:prediction_curves} 在八个代表性的 IndPenSim 异常场景上比较 GRU 主干与 GRU+ROAM。
\cref{fig:prediction_curves} compares the GRU backbone and GRU+ROAM on eight representative IndPenSim anomaly scenarios.

% 在 substrate-feed 与 aeration 异常 (A--C, G, H) 中，冻结主干会高估青霉素浓度，而 ROAM 会把预测拉回真实值。后验集中在 readout 轴，只伴随较小的 bias、load 或 scale 修正。这些偏移主要属于输出映射失配。D 和 E 更平稳。主干已经捕捉到主要批次趋势，而 ROAM 只做选择性校正。F 在八个场景中具有最大的 $\tau_r$，正向的 readout 后验仍然提供了部分补偿。H 是最困难的情形。readout 后验触及显示上限，但 ROAM 仍然只能部分缩小误差。这一结果揭示了低容量外部校正在严重 aeration mismatch 下的边界。
In the substrate-feed and aeration anomalies (A--C, G, H), the frozen backbone overestimates penicillin concentration and ROAM pulls the prediction back toward ground truth. The posterior concentrates on the readout axis with only minor bias, load, or scale corrections. These shifts are mainly output-mapping mismatch. D and E are smoother. The backbone already captures the main batch trend and ROAM makes only selective corrections. F has the largest $\tau_r$ among the eight scenarios, and a positive readout posterior still provides partial compensation. H is the hardest case. The readout posterior reaches the display limit, yet ROAM only partially closes the gap. This result exposes the boundary of low-capacity external correction under severe aeration mismatch.

\subsection{Ablation Studies}\label{sec:4-5}

% \cref{tab:ablation} 在浓密环境上对 ROAM 做消融。Var.~A 是 Main Results 中复用的冻结 GRU 基线，而 Var.~I 是完整系统。变体 B--I 共享相同的 base artifact，并依次开启标签残差更新、观测与过程证据、LLM 先验、风险门控、trust gate 和子空间解耦。
\cref{tab:ablation} ablates ROAM in the thickening environment. Var.~A is the frozen GRU baseline reused from Main Results, and Var.~I is the full system. Variants B--I share the same base artifact and progressively enable label-residual updates, observation and process evidence, LLM prior, risk gate, trust gate, and subspace decoupling.

\begin{table*}[t]
  \centering
  % 浓密环境上的组件消融（GRU 主干）。Var.~A 为冻结基线，Var.~I 为完整 ROAM。subspace 更新空间下每条证据通道只更新其指定轴，full 更新空间下所有通道更新完整 5-D latent。
  \caption{Component ablation in the thickening environment (GRU backbone). Var.~A is the frozen baseline and Var.~I is the full ROAM system. In the subspace update mode each evidence channel updates only its designated axes, whereas in the full mode all channels update the entire 5-D latent.}
  \label{tab:ablation}
  \setlength{\tabcolsep}{5pt}\scriptsize
  \begin{tabularx}{\linewidth}{c c c c c c c c rrrr}
    \toprule
            &                         & \multicolumn{3}{c}{Evidence channel} &              &              &              & \multicolumn{4}{c}{MAE (\,\texttimes\,10\textsuperscript{3}) $\downarrow$}                                             \\
    \cmidrule(lr){3-5} \cmidrule(lr){9-12}
    Variant & \makecell{Prior source} & \makecell{Label                                                                                                                                                                                            \\residual} & \makecell{Observation\\evidence} & \makecell{Process\\evidence} & \makecell{Risk gate} & \makecell{Trust gate $\tau_r$} & \makecell{Update space} & \makecell{Same-family} & \makecell{Visible shift} & \makecell{Hidden shift} & \makecell{Observation shift} \\
    \midrule
    A       & --                      & --                                   & --           & --           & --           & --                                                                         & subspace & 0.027 & 0.672 & 22.957 & 2.341 \\
    B       & --                      & $\checkmark$                         & --           & --           & $\checkmark$ & $\checkmark$                                                               & subspace & 0.026 & 0.661 & 20.461 & 2.197 \\
    C       & --                      & $\checkmark$                         & $\checkmark$ & --           & $\checkmark$ & $\checkmark$                                                               & subspace & 0.026 & 0.583 & 20.104 & 2.121 \\
    D       & --                      & $\checkmark$                         & --           & $\checkmark$ & $\checkmark$ & $\checkmark$                                                               & subspace & 0.025 & 0.638 & 19.077 & 2.003 \\
    E       & flat                    & $\checkmark$                         & $\checkmark$ & $\checkmark$ & $\checkmark$ & $\checkmark$                                                               & subspace & 0.024 & 0.556 & 18.468 & 1.862 \\
    F       & LLM                     & $\checkmark$                         & $\checkmark$ & $\checkmark$ & $\checkmark$ & --                                                                         & subspace & 0.024 & 0.552 & 18.344 & 1.841 \\
    G       & LLM                     & $\checkmark$                         & $\checkmark$ & $\checkmark$ & --           & --                                                                         & subspace & 0.024 & 0.731 & \textbf{5.306}  & \textbf{1.725} \\
    H       & LLM                     & $\checkmark$                         & $\checkmark$ & $\checkmark$ & $\checkmark$ & $\checkmark$                                                               & full     & 0.026 & 0.604 & 17.795 & 1.975 \\
    I       & LLM                     & $\checkmark$                         & $\checkmark$ & $\checkmark$ & $\checkmark$ & $\checkmark$                                                               & subspace & \textbf{0.023} & \textbf{0.551} & 18.329 & 1.752 \\
    \bottomrule
  \end{tabularx}
\end{table*}

% A$\to$B 表明在线后验更新是增益的主要来源，并且在 hidden-shift 场景上的改善最大。B$\to$C 主要改善 visible shift，而 B$\to$D 主要改善 hidden shift 和 observation shift，这证实两条通道作用于分离的 latent 子空间。观测证据校正 bias/scale 轴，而过程证据校正 load/dynamics/readout 轴。D$\to$E 在 flat prior 下补齐了三条通道，并在各个类别上都带来稳定改善。E$\to$I 则用 LLM semantics 替换了 flat prior，主要为 hidden axes 提供方向性初始化。其余列变化很小，因为标签残差和证据通道已经提供了大部分适配收益。
A$\to$B shows that online posterior updating is the primary source of gain, with the largest improvement on hidden-shift scenarios. B$\to$C mainly improves visible shifts while B$\to$D mainly improves hidden and observation shifts, confirming that the two channels act on separated latent subspaces. Observation evidence corrects bias/scale axes and process evidence corrects load/dynamics/readout axes. D$\to$E completes all three channels under a flat prior and yields consistent improvement across categories. E$\to$I replaces the flat prior with LLM semantics and mainly provides directional initialization for the hidden axes. The remaining columns change little because label-residual and evidence channels already deliver most of the adaptation benefit.

% F 和 I 仅在 trust gate 上不同。移除它会带来边际改善，这说明该门控是一种保守性机制，它以略微的精度代价来抑制不可靠的观测更新。G 进一步移除了 risk gate，在 same-family、hidden shift 和 observation shift 上取得最优 MAE，但会导致 visible shift 过度校正。H 和 I 仅在更新空间上不同。全空间更新改善了 hidden shift 的恢复，但由于跨轴干扰而抬高了其他位置的误差。子空间解耦会让每个轴的校正保持隔离。门控设计和子空间选择都用逐类别的激进性，换取跨场景的一致平衡。
F and I differ only in the trust gate. Removing it yields marginal improvement, indicating that the gate is a conservatism mechanism that suppresses unreliable observation updates at a slight accuracy cost. G further removes the risk gate and achieves the best MAE on same-family, hidden, and observation shifts, but causes visible-shift over-correction. H and I differ only in update space. Full-space updating improves hidden-shift recovery but raises error elsewhere due to cross-axis interference. Subspace decoupling keeps each axis's correction isolated. Both gating and subspace choices trade per-category aggressiveness for cross-scenario balance.

% Var.~I 是最均衡的配置。它保留了 LLM 先验对 hidden axes 的方向初始化，同时避免了 G 的过补偿和 H 的跨轴干扰。三次 LLM 重复试验在四列上的标准差都低于 0.050。
Var.~I is the most balanced configuration. It retains the LLM prior's hidden-axis initialization while avoiding the over-correction of G and the cross-axis interference of H. Standard deviations across three LLM repeats remain below 0.050.

\subsection{Analysis and Discussion}\label{sec:4-6}

% 我们分析参数敏感性、LLM 敏感性、适配范式对比和计算效率。
We examine parameter sensitivity, LLM sensitivity, adaptation paradigm comparison, and computational efficiency.

\subsubsection{Parameter Sensitivity}

\begin{figure}[t]
  \centering
  \includegraphics[width=\linewidth]{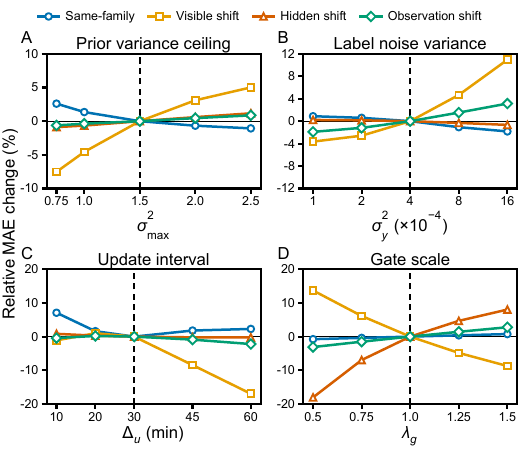}
  % GRU+ROAM 在浓密环境上四个超参数（$\sigma_{\max}^2$、$\sigma_y^2$、$\Delta_u$、$\lambda_g$）的单因素敏感性分析。每个子图报告四类场景的 MAE 随参数变化趋势。
  \caption{One-at-a-time sensitivity analysis of four ROAM hyperparameters ($\sigma_{\max}^2$, $\sigma_y^2$, $\Delta_u$, $\lambda_g$) in the thickening environment (GRU backbone). Each subplot reports MAE across four scenario categories.}
  \label{fig:sensitivity}
\end{figure}

% \cref{fig:sensitivity} 对浓密环境中 GRU+ROAM 配置的四个超参数进行单因素敏感性分析。
\cref{fig:sensitivity} presents a one-at-a-time sensitivity analysis of four hyperparameters for the GRU+ROAM configuration in the thickening environment.

% 四个超参数都表现出跨类别权衡。$\sigma_{\max}^2$ 和 $\sigma_y^2$ 的影响较温和，各偏移类别之间此消彼长。$\Delta_u$ 和 $\lambda_g$ 的影响更大。过短的更新间隔会放大短时波动，并抬高 same-family MAE。$\lambda_g$ 的权衡最明显；放松门控会显著改善 hidden shift，但会恶化 visible shift，而收紧门控则相反。默认值在四类场景之间给出了最好的折中。
All four hyperparameters show cross-category trade-offs. $\sigma_{\max}^2$ and $\sigma_y^2$ have moderate effects, with each shift category improving at the expense of others. $\Delta_u$ and $\lambda_g$ have larger effects. A short update interval amplifies short-term fluctuations and raises same-family MAE. The trade-off is most pronounced in $\lambda_g$. Relaxing the gate substantially improves hidden shift but worsens visible shift, and tightening reverses the trend. The defaults yield the best compromise across all four categories.

\subsubsection{LLM Sensitivity}

\begin{figure}[t]
  \centering
  \includegraphics[width=88mm]{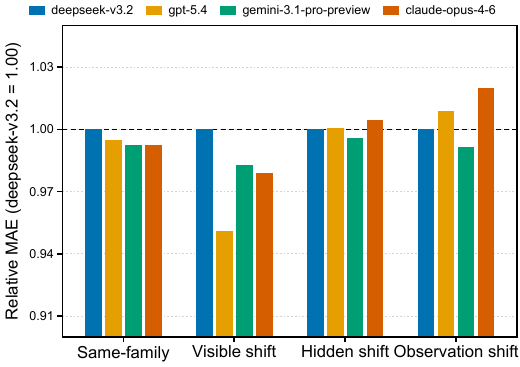}
  % LLM 先验引擎在浓密环境 GRU 主干上的相对 MAE。每类偏移场景以 DeepSeek-V3.2 的结果归一化，虚线标记 1.00 等价线。
  \caption{Relative MAE of LLM prior engines in the thickening environment with a fixed GRU backbone. Each shift category is normalized by the DeepSeek-V3.2 result. The dashed line marks parity at 1.00.}
  \label{fig:llm_sensitivity}
\end{figure}

% 为了分析 ROAM 对不同 LLM 的敏感性，\cref{fig:llm_sensitivity} 固定 GRU 主干，仅替换先验引擎，对比 DeepSeek、GPT、Gemini 和 Claude 四种 LLM 的适配效果。
To analyze the sensitivity of ROAM to different LLMs, \cref{fig:llm_sensitivity} fixes the GRU backbone and swaps only the prior engine across DeepSeek, GPT, Gemini, and Claude.

% 所有柱高都落在 0.95 到 1.05 的窄带内，四类偏移场景的最大偏差不超过 5\%。这是因为 LLM 只提供初始先验。后续在线更新由观测主导，而不同 LLM 之间的初始差异会在几个步骤后消失。结果表明，ROAM 在四种代表性 LLM 上都保持了稳定的适配性能。
All bars fall within a narrow 0.95 to 1.05 band, and the largest deviation across four shift categories does not exceed 5\%. This is because the LLM only provides the initial prior. Subsequent online updates are driven by observations, and the initial differences among LLMs vanish within a few steps. The results confirm that ROAM maintains stable adaptation performance across all four representative LLMs.
\subsubsection{Adaptation Paradigm Comparison}

% 场景变化是迁移学习、元学习和在线微调的经典适用范围。\cref{tab:paradigm} 在浓密环境上将 ROAM 与这些已有范式进行对比。
Scenario shifts fall within the classic scope of transfer learning, meta-learning, and online fine-tuning. \cref{tab:paradigm} benchmarks ROAM against these established paradigms in the thickening environment.

\begin{table}[t]
  \centering
  % 浓密环境上的适配范式对比（GRU 主干）。MAE 以 10^{-3} 为单位；各列最优值加粗。
  \caption{Adaptation paradigm comparison in the thickening environment (GRU backbone). MAE in units of \texttimes\,10\textsuperscript{3}; best per column in bold.}
  \label{tab:paradigm}
  \setlength{\tabcolsep}{4pt}\scriptsize
  \begin{tabularx}{\linewidth}{l cccc}
    \toprule
    Strategy             & \makecell{Same-family} & \makecell{Visible shift} & \makecell{Hidden shift} & \makecell{Observation shift} \\
    \midrule
    Frozen backbone      & 0.027                  & 0.672                    & 22.957                  & 2.341                        \\
    Last-layer fine-tune & 0.023                  & 0.549                    & 22.566                  & 1.983                        \\
    Full-model fine-tune & 0.026                  & \textbf{0.475}           & 22.903                  & 2.289                        \\
    MAML                 & \textbf{0.022}         & 0.510                    & 21.447                  & 1.960                        \\
    ROAM                 & 0.023                  & 0.551                    & \textbf{18.329}         & \textbf{1.752}               \\
    \bottomrule
  \end{tabularx}
\end{table}

% 两种 fine-tuning 变体都改善了 visible-shift 和 observation-shift MAE，但 hidden-shift MAE 仍然接近冻结主干的水平。这说明梯度更新能够利用可见的分布变化，却无法捕捉不可观测的过程逻辑偏移。MAML 比两种 fine-tuning 变体更进一步缩小了 hidden-shift 差距，但距离 ROAM 仍然很远，这表明当底层过程逻辑发生变化时，纯数据驱动的元学习仍然不足以补偿这种开放场景偏移。
Both fine-tuning variants improve visible- and observation-shift MAE but leave hidden-shift MAE near the frozen-backbone level. This confirms that gradient updates exploit observable distributional changes but cannot capture unobserved process-logic shifts. MAML narrows the hidden-shift gap further than either fine-tuning variant but still falls well short of ROAM, indicating that data-driven meta-learning alone cannot compensate when the underlying process logic changes.

% ROAM 在不更新主干参数的条件下取得了最优的 hidden-shift 和 observation-shift MAE，而且不会降低 same-family 性能，而 full-model fine-tuning 只在 visible shift 上领先。开放式文本证据提供了梯度信号无法获得的场景级信息，使完全冻结的主干也能实现最优的综合适配。
ROAM achieves the best hidden-shift and observation-shift MAE without updating backbone parameters or degrading same-family performance, while full-model fine-tuning leads only on visible shift. Open-ended textual evidence supplies scenario-level information that gradient signals cannot access, enabling the best overall adaptation with a fully frozen backbone.

\subsubsection{Computational Efficiency}

% \cref{tab:efficiency} 报告浓密环境上的逐步部署成本。每个 episode 仅在起始时调用一次语言模型构建语义先验。
\cref{tab:efficiency} reports per-step deployment cost in the thickening environment. Each episode invokes the language model only once at the start to construct the semantic prior.

\begin{table}[t]
  \centering
  % 浓密环境上的逐步部署成本。Peak RSS 为相对于基模型的内存增量。SVR 与 XGBoost 的主干参数包含所有存储的支持向量或树节点。
  \caption{Per-step deployment cost in the thickening environment. Peak RSS is the memory increment over the base model. For SVR and XGBoost, backbone parameters include all stored support vectors or tree nodes.}
  \label{tab:efficiency}
  \setlength{\tabcolsep}{1pt}\scriptsize
  \begin{tabularx}{\linewidth}{l rrrrr}
    \toprule
                & \multicolumn{2}{c}{Parameters (\,\texttimes\,10\textsuperscript{3}) $\downarrow$} & \multicolumn{3}{c}{Deployment Cost $\downarrow$}                              \\
    \cmidrule(lr){2-3}\cmidrule(lr){4-6}
    Backbone
                & \makecell{Backbone}
                & \makecell{Adapter}
                & \makecell{Base Inference                                                                                                                                          \\Time (ms/step)}
                & \makecell{+ROAM Inference                                                                                                                                         \\Time (ms/step)}
                & \makecell{Peak RSS (MB)}                                                                                                                                          \\
    \midrule
    SVR         & 73.489                                                                            & 0.263                                            & 0.144 & 0.151 & 0.130      \\
    XGBoost     & 7.745                                                                             & 0.263                                            & 0.255 & 0.270 & ${<}$0.001 \\
    GRU         & 0.369                                                                             & 0.167                                            & 1.532 & 1.538 & ${<}$0.001 \\
    LSTM        & 0.489                                                                             & 0.167                                            & 0.584 & 0.589 & ${<}$0.001 \\
    Transformer & 86.721                                                                            & 0.839                                            & 1.170 & 1.186 & ${<}$0.001 \\
    Informer    & 170.433                                                                           & 0.839                                            & 7.845 & 7.858 & 0.542      \\
    Mamba       & 3283.969                                                                          & 0.839                                            & 8.716 & 8.736 & ${<}$0.001 \\
    \bottomrule
  \end{tabularx}
\end{table}

% 该 adapter 最多只增加 839 个参数。在浓密环境上，所有主干的每步推理时间增量都低于 0.02\,ms，而且不需要额外的前向传播或梯度计算。七个主干中有五个的 Peak RSS 增量低于 0.001\,MB，而 SVR 和 Informer 也分别只增加了 0.130 和 0.542\,MB。相对于毫秒级的工业采样周期，总体开销可以忽略不计。
The adapter adds at most 839 parameters. In the thickening environment, per-step inference time increases by under 0.02\,ms across all backbones, requiring no additional forward pass or gradient computation. Peak RSS increment stays below 0.001\,MB for five of seven backbones, while SVR and Informer add only 0.130 and 0.542\,MB. Relative to millisecond-level industrial sampling intervals, the total overhead is negligible.

\section{Conclusion}\label{sec:5}

% 本文提出 ROAM，一种利用 LLM 世界知识与推理能力使冻结专用模型无需重训练即可适应未见场景的框架。ROAM 不修改任何主干参数，而是将 LLM 生成的场景判断与在线观测融合在低维语义校正空间中，使已部署模型能够响应新工况。当证据不足或 LLM 输出不可靠时，风险约束机制会抑制校正并回退到原始模型预测。该机制提供经验性的风险控制，而不是逐场景逐主干的单调改进保证。在矿物浓密过程和 IndPenSim 青霉素发酵数据集上的实验表明，ROAM 在主要偏移场景上显著降低误差，尤其是在隐藏偏移场景中 MAE 降低超过 20\%。整个适配仅需 839 个额外参数和不到 0.02\,ms 的每步开销。主干参数完全冻结，现有模型资产无需重建。适配过程中的每一步判断和校正都可记录和审查，便于工业现场追溯。
This article proposes ROAM, a framework that uses LLM world knowledge and reasoning to adapt frozen specialist models to unseen scenarios without retraining. ROAM keeps all backbone parameters frozen and fuses LLM-generated scenario judgments with online observations in a low-dimensional semantic correction space. This allows deployed models to respond to new operating conditions. When evidence is insufficient or LLM outputs are unreliable, a risk-constrained mechanism suppresses corrections and reverts the system to the original model prediction. This mechanism provides empirical risk control rather than a pointwise monotonic-improvement guarantee for every backbone and scenario. Experiments on a mineral thickening process and the IndPenSim penicillin fermentation dataset show that ROAM substantially reduces error in major shift settings, especially hidden shifts where MAE decreases by over 20\%. The entire adaptation requires only 839 additional parameters and under 0.02\,ms per-step overhead. Existing model assets need not be rebuilt because backbone parameters stay frozen. Every judgment and correction during adaptation is logged and auditable for field-level traceability.

% 当前框架在多种偏移来源同时剧烈变化时，校正效果会下降。当现场缺少文本形式的场景描述时，LLM 先验的信息量有限，这仍然是跨行业推广的一个瓶颈。未来工作将探索融合图像等多模态证据以减少对文本日志的依赖。我们还将引入非线性语义映射来应对更复杂的偏移模式，并将该框架扩展到其他冶金工序以验证更广泛的工业适用性。
Correction effectiveness decreases when multiple shift sources change drastically at the same time. The LLM prior carries limited information when textual scenario descriptions are unavailable on site, and this remains a bottleneck for cross-industry deployment. Future work will explore fusing multimodal evidence such as images to reduce reliance on textual logs. We will also introduce nonlinear semantic mappings for more complex shift patterns and extend the framework to other metallurgical stages for broader industrial validation.

\bibliographystyle{IEEEtran}
\bibliography{references}

\end{document}